\documentclass[sigconf]{acmart}

\usepackage{graphicx}
\usepackage{enumitem}
\usepackage{bm}
\usepackage{mathtools}
\usepackage{url}
\usepackage{algorithm}
\usepackage{algpseudocode}
\usepackage{amsmath}
\usepackage{xcolor}
\usepackage{amsthm}
\usepackage{soul}
\usepackage{bbm}
\usepackage{caption}
\usepackage{subcaption}
\usepackage{booktabs}

\newcommand{\revise}[1]{{{#1}}}

\DeclareMathOperator{\prob}{\mathbb{P}}
\def\vphi{{\bm{\phi}}}
\algrenewcommand\algorithmicrequire{\textbf{Input:}}
\algrenewcommand\algorithmicensure{\textbf{Output:}}

\usepackage{amsmath,amsfonts,bm}

\def\eqref#1{equation~\ref{#1}}
\def\1{\bm{1}}

\def\rb{{\textnormal{b}}}

\def\rt{{\textnormal{t}}}

\def\rz{{\textnormal{z}}}

\def\rva{{\mathbf{a}}}

\def\rve{{\mathbf{e}}}

\def\rvg{{\mathbf{g}}}
\def\rvh{{\mathbf{h}}}

\def\rvx{{\mathbf{x}}}
\def\rvy{{\mathbf{y}}}
\def\rvz{{\mathbf{z}}}

\def\rmX{{\mathbf{X}}}
\def\rmY{{\mathbf{Y}}}

\def\vtheta{{\bm{\theta}}}
\def\vomega{{\bm{\omega}}}
\def\va{{\bm{a}}}

\def\vg{{\bm{g}}}

\def\vs{{\bm{s}}}

\def\vx{{\bm{x}}}
\def\vy{{\bm{y}}}
\def\vz{{\bm{z}}}

\DeclareMathAlphabet{\mathsfit}{\encodingdefault}{\sfdefault}{m}{sl}
\SetMathAlphabet{\mathsfit}{bold}{\encodingdefault}{\sfdefault}{bx}{n}

\def\gA{{\mathcal{A}}}
\def\gB{{\mathcal{B}}}
\def\gC{{\mathcal{C}}}
\def\gD{{\mathcal{D}}}

\def\gG{{\mathcal{G}}}
\def\gH{{\mathcal{H}}}

\def\gL{{\mathcal{L}}}
\def\gM{{\mathcal{M}}}
\def\gN{{\mathcal{N}}}

\def\gR{{\mathcal{R}}}

\def\gT{{\mathcal{T}}}

\def\gX{{\mathcal{X}}}

\newcommand{\E}{\mathbb{E}}

\DeclareMathOperator*{\argmax}{arg\,max}

\begin{document}

\title[Analyzing Inference Privacy Risks Through Gradients In Machine Learning]{Analyzing Inference Privacy Risks Through Gradients \\ In Machine Learning}

\author{Zhuohang Li}
\affiliation{%
  \institution{Vanderbilt University}
  \city{Nashville}
  \state{TN}
  \country{USA}
  }
\email{zhuohang.li@vanderbilt.edu}

\author{Andrew Lowy}
\affiliation{%
  \institution{University of Wisconsin–Madison}
  \city{Madison}
  \state{WI}
  \country{USA}
  }
\email{alowy@wisc.edu}

\author{Jing Liu}
\affiliation{%
  \institution{Mitsubishi Electric Research Laboratories}
  \city{Cambridge}
  \state{MA}
  \country{USA}
  }
\email{jiliu@merl.com}

\author{Toshiaki Koike-Akino}
\affiliation{%
  \institution{Mitsubishi Electric Research Laboratories}
  \city{Cambridge}
  \state{MA}
  \country{USA}
  }
\email{koike@merl.com}

\author{Kieran Parsons}
\affiliation{%
  \institution{Mitsubishi Electric Research Laboratories}
  \city{Cambridge}
  \state{MA}
  \country{USA}
  }
\email{parsons@merl.com}

\author{Bradley Malin}
\affiliation{%
  \institution{Vanderbilt University}
  \city{Nashville}
  \state{TN}
  \country{USA}
  }
\email{b.malin@vanderbilt.edu}

\author{Ye Wang}
\affiliation{%
  \institution{Mitsubishi Electric Research Laboratories}
  \city{Cambridge}
  \state{MA}
  \country{USA}
  }
\email{yewang@merl.com}

\def\authors{Zhuohang, Andrew Lowy, Jing Liu, Toshiaki Koike-Akino, Kieran Parsons, Bradley Malin, Ye Wang}

\renewcommand{\shortauthors}{Li, et al.}

\begin{abstract}
In distributed learning settings, models are iteratively updated with shared gradients computed from potentially sensitive user data. While previous work has studied various privacy risks of sharing gradients, our paper aims to provide a systematic approach to analyze private information leakage from gradients. We present a unified game-based framework that encompasses a broad range of attacks including attribute, property, distributional, and user disclosures. We investigate how different uncertainties of the adversary affect their inferential power via extensive experiments on five datasets across various data modalities. Our results demonstrate the inefficacy of solely relying on data aggregation to achieve privacy against inference attacks in distributed learning.
We further evaluate five types of defenses, namely, gradient pruning, signed gradient descent, adversarial perturbations, variational information bottleneck, and differential privacy, under both static and adaptive adversary settings. We provide an information-theoretic view for analyzing the effectiveness of these defenses against inference from gradients. Finally, we introduce a method for auditing attribute inference privacy, improving the empirical estimation of worst-case privacy through crafting adversarial canary records.
\end{abstract}

\maketitle

\section{Introduction}
Ensuring privacy is an important prerequisite for adopting machine learning (ML) algorithms in critical domains that require training on sensitive user data, such as medical records, personal financial information, private images, and speech.
Prominent ML models, ranging from compact neural networks tailored for mobile platforms~\cite{howard2017mobilenets} to large foundation models~\cite{brown2020language,rombach2022high}, are
often trained on user data via gradient-based iterative optimization.
In many cases, such as decentralized learning~\cite{dhasade2023decentralized,hsieh2017gaia} or federated learning (FL)~\cite{mcmahan2017communication,hard2018federated,guliani2021training}, model gradients are directly exchanged in place of raw training data to facilitate
joint learning, which opens up an additional channel for potential privacy leakage~\cite{lowy2022private}.

Recent works have explored information leakage through this gradient channel in various forms, albeit in isolation.
For instance, Nasr \textit{et al.}~\cite{nasr2019comprehensive} showed that it is feasible to infer membership (i.e., single-bit information indicating the existence of a target record in the training data pool) from model updates in federated learning.
Beyond membership, Melis \textit{et al.}~\cite{melis2019exploiting} demonstrated inference over sensitive properties of the training data in collaborative learning.
Other independent lines of work additionally explored attribute inference~\cite{lyu2021novel,driouich2022novel} and data reconstruction~\cite{zhu2019deep,geiping2020inverting,gupta2022recovering} through shared model gradients.
However, some emerging privacy concerns that have so far only been considered under the centralized learning setting, such as the distributional inference~\cite{suri2022formalizing,chaudhari2023snap} and user-level inference~\cite{kandpal2023user,li2022user}, have not been well investigated in the gradient leakage setting.

Existing studies on information leakage from gradients have several limitations.
First, the majority of the current literature focuses on investigating each individual type of inference attack under their specific threat models while lacking a comprehensive examination of inference attack performance under various adversarial assumptions, which is essential for providing a holistic view of the adversary's capabilities.
For instance, from the attack's perspective, assuming the adversary to have access to a reasonably-sized shadow dataset and limited rounds of access to the model's gradients helps to capture the realistic inference privacy risk under a practical threat model. Conversely, from the defense's perspective, assuming a powerful adversary with access to record-level gradients and auxiliary information about the private record helps to estimate the worst-case privacy risk, which may facilitate the design of more robust defenses.
Second, while several types of heuristic defenses have been explored by prior work, their supposed effectiveness has not been fully verified under more challenging adaptive adversary settings. Moreover, existing studies do not adequately explain why some defenses succeed in reducing the inference risk over gradients, while others fail,
which could provide important guidance on the design of more effective defenses.

\begin{figure*}[t]
    \centering
    \includegraphics[width=\linewidth]{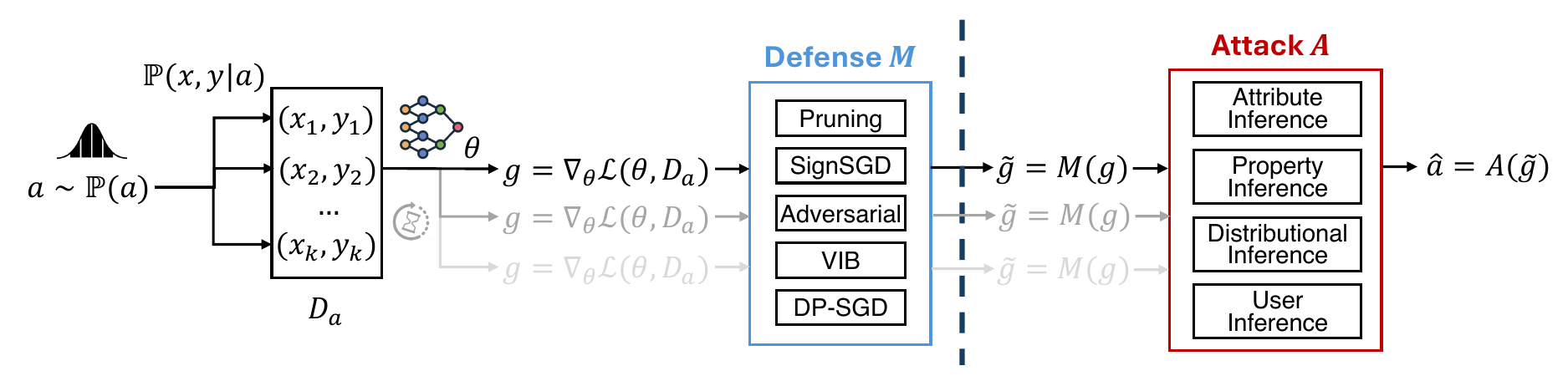}
    \caption{Overview of the unified inference game from gradients: the adversary infers the sensitive variable $\va$ from observations of the gradients $\Tilde{\vg}$ computed on the private data batch $\gD_\va$.}
    \label{fig:overview}
\end{figure*}

In this paper, we conduct a systematic analysis of private information leakage from gradients.
We start by defining a unified inference game that broadly encompasses four types of inference attacks \revise{that aims at inferring common private information of the data} from gradients, namely, \textit{attribute inference attack} (AIA), \textit{property inference attack} (PIA), \textit{distributional inference attack} (DIA), and \textit{user inference attack} (UIA), as illustrated in Figure~\ref{fig:overview}.
Under this framework, we show that information leakage from gradients can be treated as performing statistical inference over a sensitive variable upon observing samples of the gradients, with different definitions of the information encapsulated by the variable being inferred, leading to a generic template for constructing different types of inference attacks.
We additionally explore different tiers of adversarial assumptions, with varying numbers of available data samples, numbers of observable rounds of gradients, and varying batch sizes, to investigate how different priors and uncertainties in the adversary's knowledge about the gradient and data distribution affect the adversary's inferential power.

We perform a systematic evaluation of these attacks on five datasets (Adult~\cite{misc_adult_2}, Health~\cite{health_heritage}, CREMA-D~\cite{cao2014crema}, CelebA~\cite{liu2015deep}, UTKFace~\cite{zhang2017age}) with three different data modalities (tabular, speech, and image).
A common setting in distributed learning is that the data distribution is heterogeneous across different nodes but homogeneous within each node. 
\revise{Under this assumption, where the sensitive variable is common across a batch, we show that a larger batch size leads to higher inference privacy risk from gradients across all considered attacks, highlighting that \textit{solely relying on data aggregation is insufficient for achieving meaningful privacy in distributed learning.}}
With a moderate batch size (e.g., $16$), we show that an adversary can launch successful inference attacks with very few shadow data samples ($\leq1{,}000$). For instance, in the case of property inference on the Adult dataset, the adversary can achieve $0.92$ AUROC with only $100$ shadow data samples.
Moreover, we demonstrate that an adversary with access to multiple rounds of gradient updates can perform Bayesian inference to aggregate adversarial knowledge, eventually leading to higher confidence and better attack performance.

We apply the developed inference attacks to evaluate the effectiveness of five common types of defenses from the privacy literature~\cite{zhu2019deep,sun2021soteria,wu2023learning,jia2018attriguard,jia2019memguard,shan2020fawkes,song2019overlearning,scheliga2022precode,scheliga2023privacy}, including Gradient Pruning~\cite{zhu2019deep}, Signed Stochastic Gradient Descent (SignSGD)~\cite{bernstein2018signsgd}, Adversarial Perturbations~\cite{madry2018towards}, Variational Information Bottleneck (VIB)~\cite{alemi2016deep}, and Differential Privacy (DP-SGD)~\cite{abadi2016deep}, against both \textit{static} adversaries that are unaware of the defense and \textit{adaptive} adversaries that can \revise{adapt to} the defense mechanism. We find that most heuristic defense methods only offer a weak notion of ``security through obscurity'', in the sense that they defend against static adversaries empirically but can be easily bypassed by adaptive adversaries.
Although DP-SGD shows consistent performance against both static and adaptive adversaries, to fully prevent inference attacks, it often requires injecting too much noise which diminishes the utility of the learning model.
We provide an information-theoretic perspective for explaining and analyzing the (in)effectiveness of these considered defenses and show that \textit{the key ingredient of a successful defense is to effectively reduce the mutual information between the released gradients and the sensitive variable}, which could serve as a guideline for designing future defenses.
Finally, to provide practical guidance in selecting privacy parameters, we introduce an auditing approach for empirically estimating the privacy loss of attribute inference attacks through crafting adversarial canary records \revise{to approximate the privacy risk in the worst case}.

In summary, our main contributions are as follows:
\begin{itemize}[leftmargin=*]
    \item We provide a holistic analysis of inference privacy from gradients through a unified inference game that broadly encompasses a range of attacks concerning attribute, property, distributional, and user inference.

    \item \revise{We demonstrate the weakness of solely relying on data aggregation to achieve privacy against inference attacks in distributed learning. We do this through a systematic evaluation of the four types of attacks on datasets with different modalities under various adversarial assumptions.}

    \item \revise{Our analyses reveal that reducing the mutual information between the released gradients and the sensitive variable is the key ingredient of a successful defense. This is shown by investigating five common types of defense strategies against inference over gradients from an information-theoretic perspective.}

    \item \revise{Our auditing results provide an empirical justification for tolerating large DP parameters when defending against attribute inference attacks (c.f.~\cite{lowy2024does}). This is achieved by implementing an auditing method for empirically estimating the privacy loss against attribute inference attacks from gradients.}

\end{itemize}

\section{Background and Related Work}

\subsection{Machine Learning Notation}
A machine learning (ML) model can be denoted as a function $f_\vtheta: \rvx \rightarrow \rvy$ parameterized by $\vtheta$ that maps from the input (feature) space to the output (label) space.
The training of an ML model involves a set of training data and an optimization procedure, such as stochastic gradient descent (SGD). At each step of SGD, a loss function $\mathcal{L}(\vtheta, \gD_b)$ is first computed based on the current model and a batch of $k$ training samples $\gD_b = \{ (\vx_i, \vy_i) \}_{i=1}^k$ and then a set of gradients is computed as $\vg = \nabla_{\vtheta} \mathcal{L}(\vtheta, \gD_b)$. Finally, the model is updated by taking a gradient step towards minimizing the loss.

\subsection{Related Work}

Developing ML models in many applications involves training on the users' private data, which introduces privacy leakage risks from different components of the ML model across several stages of the development and deployment pipeline.

\textbf{Leakage From Model Parameters ($\vtheta$).}
The first way of exposing privacy information is through analyzing the model parameters.
This is connected to the most prominent centralized ML setting, where the model is first developed on a local dataset and then released to the users for deployment.
Various forms of privacy leakage have been studied in this setting. 
White-box membership inference~\cite{leino2020stolen,nasr2019comprehensive,sablayrolles2019white} aims at identifying the presence of individual records in the training dataset given access to the full model.
Data extraction attacks exploit the memorization of the ML model to extract training samples~\cite{haim2022reconstructing,carlini2023extracting}, whereas model inversion attacks generate synthetic data samples from the training distribution~\cite{yin2020dreaming,wang2021variational}.
In contrast, for distributional inference attacks~\cite{ateniese2015hacking,ganju2018property,suri2022formalizing}, the attacker's goal is to make inferences about the entire training data distribution rather than individuals.

\textbf{Leakage From Model Outputs ($f_\vtheta(\vx)$).}
Another source of privacy leakage is the model output, which is related to more restrictive settings such as machine learning as a service (MLaaS) in cloud APIs where only black-box access to the ML model is granted. Under this setting, researchers have studied several privacy attacks that can be launched by querying the model and observing the outputs.
For instance, query-based model inversion attacks~\cite{fredrikson2014privacy,fredrikson2015model} exploit the predicted confidence or labels from the model to make inferences about the input data instance~\cite{zhang2020secret} or attribute~\cite{mehnaz2022your}.
Model stealing attacks attempt to recover the confidential model weights~\cite{tramer2016stealing} or hyper-parameters~\cite{wang2018stealing} given query access to the model.
Black-box membership inference attacks~\cite{salem2018ml,truex2019demystifying,sablayrolles2019white,song2021systematic} and black-box distributional inference attacks~\cite{mahloujifar2022property,chaudhari2023snap} allow an adversary to decide whether a data point was included in training or reveal information about the training data distribution by analyzing its output prediction or confidence.

\textbf{Leakage From Model Gradients ($\vg$).}
The final source of privacy leakage is the gradient of the loss function with respect to the model parameters, which is essential for updating the model with stochastic gradient descent. This is relevant to ML settings that release intermediate model updates during model development, such as distributed training, federated learning, peer-to-peer learning, and online learning.
Compared to model parameters, model gradients carry more nuanced information about a small batch of data used for computing the update and thus may reveal more information about the underlying data instances.
Current literature studies different types of gradient-based privacy leakage in isolation.
One line of work focused on data reconstruction from model gradients~\cite{zhu2019deep,geiping2020inverting} or updates~\cite{salem2020updates,haim2022reconstructing} with various data types, such as image~\cite{zhu2019deep,geiping2020inverting,yin2021see,li2022auditing}, text~\cite{gupta2022recovering,haim2022reconstructing}, tabular~\cite{vero2023tableak}, and speech data~\cite{li2023speech}.
However, these attacks rely on strong adversarial assumptions and do not generalize to large batch sizes~\cite{huang2021evaluating}.
Another line of work investigated the extraction of private attributes or properties~\cite{melis2019exploiting,feng2021attribute} of the private data from model gradients.
Specifically, Melis \textit{et al.}~\cite{melis2019exploiting} first revealed that gradients shared in collaborative learning can be used to infer properties of the training data that are uncorrelated with the task label.
Lyu \textit{et al.}~\cite{lyu2021novel} explored attribute reconstruction from epoch-averaged gradients on tabular and genomics data.
Feng \textit{et al.}~\cite{feng2021attribute} discovered that gradients of Speech Emotion Recognition models leak information about user demographics such as gender and age.
Dang \textit{et al.}~\cite{dang2022method} showed that speaker identities can be revealed from the gradients of Automatic Speech Recognition models.
\revise{Kerkouche \textit{et al.}~\cite{kerkouche2023client} demonstrated the weakness of secure aggregation without differential privacy in Federated learning by designing a disaggregation attack that exploits the linearity of model aggregation and client participation across multiple rounds to capture client-specific properties.}
In contrast to existing studies that design separate treatments for each type of attack, in this work, we take a holistic view of information leakage from gradients.

\section{Problem Formalization}

This section introduces four types of inference attacks from gradients, namely, \textit{attribute inference}, \textit{property inference}, \textit{distributional inference}, and \textit{user inference}. We formally define information leakage from gradients using a unified security game, following standard practices in machine learning privacy studies~\cite{salem2023sok}, and discuss variants of threat models that affect the adversary's inferential power. In Section~\ref{sec:atk_construction}, we describe methods to construct these attacks.

\subsection{Attack Definitions}

We consider four types of information leakage from model gradients that generally involve two parties, namely, a private learner who releases model gradients computed on a private data batch, and an adversary who tries to make inferences about the private data given access to the gradients.
This generic setting captures multiple ML application scenarios such as distributed training, federated learning, and online learning.

\textbf{Attribute Inference.}
Attribute inference attacks (AIA) seek to infer a data record's unknown attribute (feature) from its gradient.
Prior works in both centralized~\cite{wu2016methodology,yeom2018privacy} and federated settings~\cite{lyu2021novel,driouich2022novel} usually assume the record to be \textit{partially known}.
For instance, infer a missing entry (e.g., genotype) of a person's medical record~\cite{fredrikson2014privacy}.
It is worth noting that, in practice, when the attributes are not completely independent, an adversary with partial knowledge about the record may be able to infer the unknown attribute just from the known ones, as in data imputation~\cite{jayaraman2022attribute}.

\textbf{Property Inference.}
Property inference attacks (PIA) aim to infer a global property of the private data batch that is not directly present in the data feature space but is correlated with some of the features (and consequently the gradients). For tabular data, these properties could be sensitive features that have been intentionally excluded from training (e.g., pseudo-identifiers in health records that are required to be removed for HIPAA compliance); for high-dimensional data like image and speech, they could be some high-level statistical features capturing the semantics of the data instance (e.g., race of a face image~\cite{melis2019exploiting} or gender of a speech recording~\cite{feng2021attribute}).

\textbf{Distributional Inference.}
Distributional inference attacks (DIA) aim to infer the ratio of the training samples ($\alpha$) that satisfy some target property\footnote{Some prior work also refers to distributional inference as property inference.}.
The majority of current literature on DIA~\cite{ganju2018property,suri2022formalizing,mahloujifar2022property,chaudhari2023snap} is in the space of centralized learning, which captures leakage from model parameters. These studies usually define DIA as a distinguishing test between two worlds where the model is trained on two datasets with different ratios ($\alpha_0$ and $\alpha_1$)~\cite{suri2022formalizing}. This can be further categorized into \textit{property existence} tests that decide if there exists any data point with the target property in the training set and \textit{property size estimation} tests that infer the exact ratio of the property in the training data~\cite{chaudhari2023snap}.
In this work, we extend DIA to the gradient space and consider a general case that combines property existence and property size estimation
by formulating DIA as performing ordinal classification between a set of $m$ ratio bins ($m\geq3$), i.e., $\{0\}, (0,\frac{1}{m-1}], (\frac{1}{m-1},\frac{2}{m-1}], ..., (\frac{m-2}{m-1}, 1]$.

\textbf{User Inference.} User inference attacks (UIA) or re-identification attacks aim to identify which user's data was used to compute the observed gradients. Here, the adversary does not know the user's exact data used for computing the gradients. Instead, the adversary is provided a set of candidate users and their corresponding underlying user-level data distributions. This setting shares similarities with the \textit{subject-level membership inference}~\cite{suri2022subject} in the sense that both attacks measure the privacy risk at the granularity of each individual. \revise{However, the user inference attack aims to infer richer information that directly exposes the user's identity compared to the membership inference attack, which only discloses a single bit of information (i.e., whether a given user’s data sample is involved in training). Thus user inference can be considered as a generalization of subject-level membership inference attack.}

We note that except for attribute inference which directly exposes (part of) the user's private data, property inference, distributional inference, and user inference attacks are \textit{inferential disclosures} (also known as \textit{deductive disclosures}) that exploit the statistical correlation exists in data to infer sensitive information from the released gradients with high confidence.
\revise{We exclude record-level privacy attacks such as membership inference and data reconstruction as our analysis here focuses on distributed learning scenarios where private information can be shared across different data samples within a batch.}

\subsection{Unified Inference Game}\label{subsec:inference_gam}

Our framework aims to capture an abstraction of privacy problems in distributed learning settings, where an attacker aims to recover some sensitive information of a particular client from their shared gradients (or model updates).
In practical distributed learning settings, the data may be heterogeneously split across the clients, and an attacker may take advantage of side information about a particular client's local data distribution.
Generally, the objective of the attacker is to recover the sensitive information, represented by the variable $\rva$, which is related to the local data distribution of the client through a joint distribution $\prob(\rvx, \rvy, \rva) = \prob(\rva) \prob(\rvx, \rvy | \rva)$.
As we will detail later, specific choices in what $\rva$ represents and the corresponding specialized structure of $\prob(\rvx, \rvy, \rva)$ enable the framework to capture attribute, property, distributional, and user inference privacy problems.
This joint distribution may capture both the side information available to the attacker and the inherent heterogeneity of the data.
To focus on evaluating the effectiveness of gradient-based attacks and defenses, 
we simplify the modeling of the overall training procedure, by updating the model in a centralized fashion on the entire training data set $\gD$, but generating gradients for the attacker on batches drawn according to $\prob(\rvx, \rvy, \rva)$.

\definition{\textit{\textbf{Unified Inference Game.}}}\label{def:unified_game}
Let $\prob(\rvx, \rvy, \rva)$ be the joint distribution, $\gL$ the loss function, $\gT$ the training algorithm, $r$ the total number of training rounds, and $\gR \subset [r]$ a set of rounds that are observable to the adversary\footnote{We use $[r]$ to denote the discrete set $\{1, 2, ..., r\}$.}.
The unified inference game from gradients between a challenger (private learner) and an adversary is as follows:

\begin{enumerate}[label={(\arabic*)}]
    \item Challenger initializes the model parameters as $\vtheta_0$.

    \item Challenger samples a training dataset $\gD=\{ (\vx_j, \vy_j) \}_{j=1}^n$, %
    where $(\vx_j, \vy_j) \overset{\mathrm{i.i.d.}}{\sim} \prob(\rvx, \rvy)$.
    
    \item Challenger draws the sensitive variable $\va \sim \prob(\rva)$.

    \item Challenger draws a batch of $k$ data samples $\gD_\va = \{ (\vx_p, \vy_p) \}_{p=1}^k$, where $(\vx_p, \vy_p) \overset{\mathrm{i.i.d.}}{\sim} \prob(\rvx, \rvy | \va)$, for the given $\va$.

    \item\label{step:start} Challenger computes the gradient of the loss on the data batch, $\vg_i = \nabla_{\vtheta_{i-1}} \mathcal{L}(\vtheta_{i-1}, \gD_\va)$.
    
    \item Challenger applies the defense mechanism $\mathcal{M}$ to produce a privatized version of the gradient $\Tilde{\vg}_i = \mathcal{M}(\vg_i)$.
    When no defense is applied, $\mathcal{M}$ is simply the identity function, i.e., $\Tilde{\vg}_i = \vg_i$.

    \item\label{step:end} The model is updated by applying the training algorithm on the training dataset for one epoch $\vtheta_{i} \leftarrow \gT(\vtheta_{i-1}, \gD, \mathcal{L}, \mathcal{M})$.

    \item Steps \ref{step:start}-\ref{step:end} are repeated for $r$ rounds.

    \item A \textit{static} adversary $\gA_s$ gets access to $\gL$, $\gT$, $\prob(\rvx, \rvy, \rva)$, and the set of (intermediate) model parameters $\Theta=\{\vtheta_{i-1}| i \in \gR \}$ and released gradients $\gG=\{\Tilde{\vg}_i|i\in \gR \}$. An \textit{adaptive} adversary $\gA_a$ also gets the defense mechanism $\mathcal{M}$.

    \item The adversary outputs its inference $\hat{\va}$ of the sensitive variable, i.e., $\hat{\va}\leftarrow \gA_s(\gL, \gT, \prob(\rvx, \rvy, \rva), \Theta, \gG)$ for the static adversary, or $\hat{\va}\leftarrow \gA_a(\gL, \gT, \prob(\rvx, \rvy, \rva), \Theta, \gG,\mathcal{M})$ for the adaptive adversary. The adversary wins if $\hat{\va} = \va$ and loses otherwise.
\end{enumerate}

In the above general game, the flexibility of the joint distribution $\prob(\rvx, \rvy, \rva)$ allows capturing various scenarios.
Rather than explicitly defining this joint distribution, which anyways depends on the unknown data distribution, we implicitly define it through transformations/filtering of a given data set.
Further, providing the adversary with knowledge of the distribution $\prob(\rvx, \rvy, \rva)$ is realized by providing the adversary with suitable shadow datasets drawn according to such transformations and filtering operations.

\textbf{Attribute Inference Game.}
The variable $\va\in[m]$ is a discrete attribute within the features $\vx$.
Sampling $\va \sim \prob(\rva)$ is accomplished by drawing uniformly or according to its marginal empirical distribution within the given training data set $\gD$. Drawing the data batch $\gD_\va$ according to the distribution $\prob(\rvx, \rvy | \va)$, is accomplished by uniformly selecting data samples $(\vx, \vy)$ from the entire training data set $\gD$ with features $\vx$ that possess the attribute $\va$.

\textbf{Property Inference Game.}
This scenario is similar to attribute inference, except that $\va\in[m]$ is a property associated with, but external to the features of, each data sample (i.e., $\va$ may be some meta-data property of each sample, but excluded from the features of $\vx$).
Drawing the data batch $\gD_\va$ is handled similarly to the attribute inference case.

\textbf{Distributional Inference Game.} In this class of scenarios, we have a general set of $m$ transformations $\{\Phi_\va | \va \in [m]\}$, which are selected by the sensitive variable $\va$.
Each transformation $\Phi_\va$ corresponds to implicitly realizing the corresponding $\prob(\rvx, \rvy | \va)$, by applying a general transformation that involves selective sampling from the overall training set $\gD$.
For example, the selection of $\va$ may indicate a particular proportion for the prevalence of a certain attribute or property, and thus the corresponding transformation would select batches of data according to that proportion.

\textbf{User Inference Game.} This is a special case of property inference, where $\va$ specifically corresponds to the identity of an individual that provided the corresponding data samples.
Unlike other inference attacks, the sensitive variable, as it represents identity, does not take on a fixed set of values. To make the attack more operational, similar to prior work on data reconstruction~\cite{hayes2024bounding}, we assume the inference is over a fixed set of $m$ candidate users randomly sampled from the population at the beginning of each game.

\subsection{Threat Model}
In this work, we assume the adversary has no control over the training protocol and only passively observes gradients as the model is being updated.
In practice, the adversary could be an honest-but-curious parameter server~\cite{li2014scaling} in a distributed learning or federated learning setting, a node in decentralized learning~\cite{dhasade2023decentralized}, or an attacker who eavesdrops on the communication channel.
The game as defined in Definition~\ref{def:unified_game} is similar to games defined in many prior works~\cite{carlini2022membership,yeom2018privacy} which captures the average-case privacy as the performance of the attack is measured by its expected value over the random draw of data samples.
In Section~\ref{sec:privacy_audit}, we consider an alternative game where the data samples are adversarially chosen to provide a measure of worst-case privacy for privacy auditing.

We consider the following aspects that reflect different levels of the adversary's knowledge:

\begin{itemize}[leftmargin=*]
    \item \textbf{Knowledge of Data Distribution.}
Similar to many prior works on inference attacks~\cite{shokri2017membership,melis2019exploiting,ye2022enhanced,suri2022formalizing,carlini2022membership,liu2022ml,chaudhari2023snap}, we model the adversarial knowledge of the data distribution through access to data samples drawn from this distribution, which are referred to as shadow datasets. A larger shadow dataset implies a more powerful adversary that has more knowledge about the underlying data distribution.
For discrete attributes, we additionally consider a more informed adversary who knows the prior distribution of the attribute, which can be estimated by drawing a large amount of data from the population.

    \item \textbf{Continuous Observation.} We use the observable set $\gR$ to capture the adversary's ability to observe the gradients continuously. Intuitively, an adversary observing multiple rounds should perform better than a single-round adversary.
Assuming a powerful adversary is beneficial for analyzing and auditing defenses. For instance, the privacy analysis in DP-SGD~\cite{abadi2016deep} assumes that the adversary has access to all rounds of gradients.

    \item \textbf{Adaptive Adversary.} When evaluating defenses, in addition to the static adversary, we consider a stronger adaptive adversary who is aware of the underlying defense mechanism. This has been demonstrated as pivotal for thoroughly assessing the effectiveness of security defenses~\cite{carlini2017adversarial,tramer2020adaptive}.
\end{itemize}

\section{Attack Construction}\label{sec:atk_construction}

\subsection{Inference Attacks}

The objective of the inference adversary is to infer the sensitive variable from the observed gradient, i.e., modeling the posterior distribution $\prob (\rva|\rvg)$.
The general strategy of implementing inference attacks from gradients is to exploit the following two adversarial assumptions as defined in the unified inference game in Section~\ref{subsec:inference_gam}.
First, the adversary possesses knowledge about the underlying population data distribution. Operationally, this implies that the adversary is able to draw data samples $(\vx, \vy)$ with corresponding sensitive variable $\va$ from $\prob(\rvx, \rvy, \rva)$ to construct a shadow dataset.
Second, the adversary has access to the training algorithm and the current model parameters, which allows the adversary to compute the gradients $\vg$ for each batch of samples within the shadow dataset.
With this information, the adversary can train a predictive model $P_\vomega(\rva|\rvg)$ to approximate the posterior.

\textbf{Attribute \& Property Inference.}
The attribute and property inference attacks follow a similar attack procedure, with the difference being whether the sensitive variable $\rva$ is internal or external to the data record.
Specifically, the adversary first constructs a shadow dataset $\gD_\vs$ by sampling from the population distribution, i.e., $\gD_\vs = \{ (\vx_j, \vy_j, \va_j) \}_{j=1}^s$ where $(\vx_j, \vy_j, \va_j) \overset{\mathrm{i.i.d.}}{\sim} \prob(\rvx, \rvy, \va)$.
Then the adversary draws data batches $\gD_\va= \{ (\vx_j, \vy_j) \}_{j=1}^k$ from the shadow dataset through bootstrapping. This is achieved by repeatedly sampling the sensitive attribute $\va$ and then drawing $k$ records that have the sensitive attribute from $\gD_\vs$. 
Next, for each data batch $\gD_\va$, the adversary computes the gradient $\vg_\va = \nabla_{\vtheta} \mathcal{L}(\vtheta, \gD_\va)$ using the current model parameters $\vtheta$.
This results in a set of labeled data pairs $(\vg_\va, \va)$, which can then be used for training an ML model $P_\vomega(\rva|\rvg)$ that predicts the sensitive variable from gradient observations.
In practice, we find that it is beneficial to train the predictive model using a balanced dataset, which can be seen as modeling $\frac{\prob (\rva|\rvg)}{\prob (\rva)}$, and capture the prior knowledge in a separate term. This provides more stable performance for small shadow dataset sizes and skewed sensitive variable distributions.

It is worth noting that here we are considering a more restrictive setting for attribute inference where the adversary holds no additional knowledge about the private data besides the gradients compared to prior works that assume the private record to be partially known (e.g., \cite{lyu2021novel,driouich2022novel} assume that everything is known except for the sensitive attribute).
Our framework can be easily extended to the general case where the adversary holds arbitrary additional knowledge $\varphi(\vx)$ about the private record $\vx$ by training a predictive model $P_\vomega(\rva|\rvg, \varphi(\vx))$ using shadow data drawn from $\prob(\rvx, \rvy, \rva|\varphi(\vx))$.

\textbf{Distributional Inference.}
In distributional inference, the sensitive variable is the index of the ratio bin to which the property ratio belongs.
The adversary first samples a random bin index $\va$ and then samples a property ratio $\alpha$ within that bin.
Next, the adversary draws a data batch $\gD_\va$ with $\lfloor \alpha k \rfloor$ records with the property and the rest without the property and derives the gradient $\vg_\va$. This process is repeated by the adversary to collect a set of labeled gradients and attribute pairs $(\vg_\va, \va)$ to train a predictive model.
We note that in the setting of distributional inference, the sensitive variable is a series of ordinal numbers indicative of the continuous property ratio $\alpha$ and thus should not be treated as regular multi-class classification.
To utilize the ordering information, we adopt a simple strategy to ordinal classification~\cite{frank2001simple}, which transforms the $m$-class ordinal classification problem into $m-1$ binary classifications. Specifically, the adversary trains a series of $m-1$ binary classifiers, with the $i$-th classifier $P_{\vomega_i}(\rva>i|\rvg)$ trained to decide whether or not $\va$ is larger than $i$. The final posterior probability can be obtained as
\begin{align*}
    P_\vomega(\rva=\va|\rvg) =
    \begin{cases}
    1 - P_{\vomega_1}(\rva>1|\rvg),& \text{if } \va=1\\
    P_{\vomega_{\va-1}}(\rva>\va-1|\rvg) - P_{\vomega_\va}(\rva>\va|\rvg),& \text{if } 1<\va<m\\
    P_{\vomega_{m-1}}(\rva>m-1|\rvg),& \text{if } \va=m\\
\end{cases}.
\end{align*}

\textbf{User Inference.}
In contrast to other inference attacks where the sensitive variable is sampled from a well-defined set of values, in user inference, the sensitive variable is the user's identity, which does not take on a fixed set of values.
Moreover, the identities that occur during test time are likely not seen during the development of the attack model. As a result, the posterior $\prob(\rva|\rvg)$ cannot be directly modeled.
To resolve this, we employ a training strategy analogous to the prototypical network~\cite{snell2017prototypical} for few-shot learning. Specifically, we first train a neural network $f_\vomega \circ u$ that is composed of an encoder $f_\vomega: \rvg \rightarrow \rvh$ that maps the gradient vector to a continuous embedding space and a classifier $u: \rvh \rightarrow \rva$ that takes the embedding as input and outputs the predicted user identity. Given gradient and sensitive variable pairs $(\vg, \va)$ created from the shadow dataset, as the number of available users in the shadow dataset is finite, the neural network can be trained in an end-to-end manner using standard multi-class classification loss such as cross-entropy. After training, the classifier $u$ is discarded. At the time of inference, the adversary is provided with an observed gradient $\Tilde{\vg}$ and a set of $m$ candidate data batches $\{\gD_i|i\in[m]\}$, where $\gD_i=\{ (\vx_j, \vy_j) \}_{j=1}^k$.
Then, the adversary can derive the corresponding set of candidate gradients $\{\vg_i|i\in[m]\}$ based on the current model parameters $\vtheta$.
Finally, the adversary computes the probability of each candidate identity after observing the gradient as
\begin{equation*}P_\vomega(\rva=\va|\rvg=\Tilde{\vg})=\frac{\exp{(-||f_\vomega(\vg_\va) - f_\vomega(\Tilde{\vg})||_2)}}{\sum_{i \in [m]} \exp{(-||f_\vomega(\vg_i) - f_\vomega(\Tilde{\vg})||_2)}}.
\end{equation*}

\subsection{Continual Attack and Adaptive Attack}

The inference attack can be further improved if the adversary has access to multiple rounds of gradients or the defense mechanism.

\textbf{Inference under Continual Observation.}
In cases where continual observation of the gradients is allowed, the adversary can use the set of observed gradients $\gG=\{\Tilde{\vg}_i|i\in \gR\}$ from multiple rounds to improve the attack. A naive solution would be to train a model to directly approximate $\prob(\rva|\gG)$. However, this would be generally infeasible in practice because of the high dimensionality of $\gG$.
Instead, the adversary can use Bayesian updating to accumulate adversarial knowledge.
Specifically, given a set of observed gradients, the log-posterior can be formulated as
\begin{align}
    \log &\prob(\rva=\va|\gG) \\ 
    & = \log{\prob(\gG|\rva=\va)} + \log{\prob(\rva=\va)} - \log{\prob(\gG)} \label{eq:bayes} \\ 
    & \approx \sum_{i\in \gR}\log{\prob(\Tilde{\vg}_i|\rva=\va)} + \log\prob(\rva=\va) -  \log{\prob(\gG)} \label{eq:indepedence} \\ 
    & = \sum_{i\in \gR} \bigg(\log \prob(\rva=\va|\Tilde{\vg}_i) + \log\prob(\Tilde{\vg}_i) - \log\prob(\rva=\va) \bigg) \nonumber \\ & \qquad + \log\prob(\rva=\va) - \log{\prob(\gG)} \\
    & = \sum_{i\in \gR} \log\prob(\rva=\va|\Tilde{\vg}_i) - (|\gR|-1)\log\prob(\rva=\va) + \gC, \label{eq:multi-round}
\end{align}
where Eq.~(\ref{eq:indepedence}) makes the approximating assumption that the gradients are conditionally independent given $\va$.
Since $\gC = - \log{\prob(\gG)} + \sum_{i\in \gR}\log{\prob(\Tilde{\vg}_i)} $ is independent of $\va$, and therefore it can be treated as a constant. $\gC=0$ if the gradients $\Tilde{\vg}_i$ are additionally mutually independent. In Eq.~(\ref{eq:multi-round}), the prior term is known and $\prob(\rva=\va|\Tilde{\vg}_i)$ can be approximated by training a fresh model for each round of observation. The sensitive variable can thus be estimated as $\hat{\va}=\argmax_\va \log\prob(\rva=\va|\gG)$.

\textbf{Adaptive Attack.}
The adversary can design adaptive attacks if the defense mechanism $\gM$ is known.
Instead of training the predictive model $P_\vomega(\rva|\rvg)$ using clean gradient pairs $(\vg_\va, \va)$, a simple strategy for adaptive attack is to apply the same defense mechanism to the shadow data's gradients and use the transformed gradient pairs $(\gM(\vg_\va), \va)$ to train the predictive model $P_\vomega(\rva|\gM(\rvg))$.
As we will show in Section~\ref{sec:defense}, this simple strategy is sufficient to bypass several heuristic-based defenses.

\section{Attack Evaluation}

\revise{In this section, we evaluate the four inference attacks on datasets with different modalities to investigate the impact of various adversarial assumptions. The findings we present below indicate the key factors that affect the attack performance are: (1) \textit{Continual Observation}: an adversary can improve the inference by accumulating information from multiple rounds of updates, (2) \textit{Batch Size}: when the private information is shared across the batch, using a large batch averages out the effect of the other variables, making it easier to infer the sensitive variable, and (3) \textit{Adversarial Knowledge}: the attack improves with the amount of knowledge of the data distribution (as captured by the number of available shadow data points).}

\subsection{Experimental Setup}

\subsubsection{Datasets and Model Architecture.} We consider the following five datasets with different data modalities (tabular, speech, and image) in our experiments.

\begin{table}[t]
\centering
\caption{Summary of datasets used in experiments.}
\label{tab:data_summary}
\resizebox{\linewidth}{!}{
\begin{tabular}{c||c|c|c|c}
\toprule
\textbf{Dataset} & \textbf{Type} & \textbf{Task Label} & \textbf{Sensitive Variable} & \textbf{Correlation} \\ \midrule
Adult            & Tabular       & Income              & Gender                         & -0.1985              \\ 
Health           & Tabular       & Mortality            & Gender                         & -0.1123              \\ 
CREMA-D          & Speech        & Emotion             & Gender                      & -0.0133              \\ 
CelebA           & Image         & Smiling            & High Cheekbones             & 0.6904               \\ 
UTKFace         & Image         & Age                 & Ethnicity                   & -0.1788              \\ \bottomrule
\end{tabular}
}
\end{table}

\begin{enumerate}[label={(\arabic*)},leftmargin=*]
    \item \textbf{Adult}~\cite{misc_adult_2} is a tabular dataset containing $48{,}842$ records from the 1994 Census database. We train a fully-connected neural network to predict the person's annual income (whether or not more than $50$K a year) and use gender (male or female) as the private attribute. For property and distributional inference attacks, the sex feature is removed.

    \item \textbf{Health}~\cite{health_heritage} (Heritage Health Prize) is a tabular dataset from Kaggle that contains de-identified medical records of over $55{,}000$ patients' inpatient or emergency room visits. We train a fully-connected neural network to predict whether the Charlson Index (an estimate of patient mortality) is greater than zero. We use the patient's gender (male, female, or unknown) as the private attribute, which is removed for property and distributional inference attacks.

    \item \textbf{CREMA-D}~\cite{cao2014crema} is a multi-modal dataset that contains $7{,}442$ emotional speech recordings collected from $91$ actors ($48$ male and $43$ female). Speech signals are pre-processed using OpenSMILE~\cite{eyben2010opensmile} to extract a total number of $23{,}990$ utterance-level audio features for automatic emotion recognition. Following prior work~\cite{feng2021attribute}, we use EmoBase which is a standard feature set that contains the MFCC, voice quality, fundamental frequency, and other statistical features, resulting in a feature dimension of $988$ for each utterance~\cite{haider2021emotion}. We train a fully connected neural network to classify four emotions, including happy, sad, anger, and neutral. We use the speaker's gender (male or female) as the target property for inference attacks.

    \item \textbf{CelebA}~\cite{liu2015deep} contains $202{,}599$ face images, each of which is labeled with $40$ binary attributes. We resize the images to $32\times32$ pixels and train a convolutional neural network to classify whether the person is smiling and use whether or not the person has high cheekbones as the target property.

    \item \textbf{UTKFace}~\cite{zhang2017age} consists of over $20{,}000$ face images annotated with age, gender, and ethnicity. We resize the images to $32\times32$ pixels and select $22{,}012$ images from the four largest ethnicity groups (White, Black, Asian, or Indian) to train a convolutional neural network to classify three age groups ($0-30$, $31-60$, and $\geq61$ years old). Ethnicity is used as the target property.
\end{enumerate}

\noindent We split each dataset three-fold into a training set, a testing set, and a public set. The training set is considered to be private and is only used for model training and inference attack evaluation. The testing set is reserved for evaluating the utility of the ML model. The public set is accessible to both the adversary and the private learner, which can be used as the shadow dataset for training the adversary's predictive model or developing defenses as described in Section~\ref{sec:defense}.
We provide a summary of the datasets in Table~\ref{tab:data_summary}, including the task label $\rvy$, the sensitive variable $\rva$ for AIA and PIA, and the Pearson correlation between $\rvy$ and $\rva$.

\subsubsection{Metrics.}
We define the following metrics for measuring inference attack performance:
\begin{enumerate}[label={(\arabic*)},leftmargin=*]
\item \textbf{Attack Success Rate (ASR)}: We measure the attack performance by the number of times the adversary successfully guesses the sensitive variable, i.e., $p=\sum_{t\in[T]} \mathbbm{1}_{\hat{\va}=\va}/T$, where $T$ is the total number of trials (i.e., repetitions of the inference game). 

\item \textbf{AUROC}: We additionally report the area under the receiver operating characteristic curve (AUROC). For sensitive variables that have more than two classes, we report the macro-averaged AUROC.

\item \textbf{Advantage}: We follow prior work~\cite{yeom2018privacy,guo2023analyzing} and use the advantage metric to measure the gain in the adversary's inferential power upon observing the gradients.
Specifically, the advantage of an adversary is defined by comparing its success rate $p$ to a baseline adversary who doesn't observe the gradients, i.e., $\texttt{Adv}(p) \coloneqq {\max(p-p^*, 0)}/{(1-p^*)} \in [0, 1]$, where $p^*$ is the success rate of the baseline adversary.
The Bayes optimal strategy for the baseline adversary without observing gradients is to guess the majority class, i.e., $p^*=\argmax_\va \prob(\rva=\va)$.

\item \textbf{TPR@$1\%$FPR}: Besides average performance metrics, recent work on membership inference~\cite{carlini2022membership,ye2022enhanced} argue the importance of understanding the privacy risk on worst-case training data by examining the low false positive rate (FPR) region. Inspired by this, we additionally report the true positive rate (TPR) when the FPR is $1\%$.
\end{enumerate}

\subsubsection{Adversary's Model.} We conducted preliminary experiments with various types and configurations of ML models and found that random forest with $50$ estimators performs the best (especially in the low FPR region) for estimating the posterior in AIA, PIA, and DIA with small shadow dataset sizes. For UIA, we use a fully-connected network with one hidden layer as the encoder. The embedding dimension is set to be $50$ for the CREMA-D dataset of $100$ for CelebA dataset. As the gradient vector is extremely high dimensional (e.g., the gradient dimensions for CREMA-D and CelebA datasets are $67{,}716$ and $45{,}922$, respectively), we apply a $1$-dimensional max-pooling layer before the adversary's predictive model with a kernel size of $3$ for tabular datasets and $10$ for other datasets for dimensionality reduction.

\subsubsection{Other Attack Settings.}
We assume the model parameters $\vtheta$ are randomly initialized at the beginning of the inference game. During the game, the model parameters are updated at each epoch using SGD with a learning rate of $0.01$.
We evaluate AIA on the tabular datasets and UIA on datasets that contain user labels (CREMA-D and CelebA), while PIA and DIA are evaluated on all datasets.
For AIA, PIA, and DIA, we use a training set of $5{,}000$ samples and a balanced public set that contains a default number of $1{,}000$ samples equally divided for each sensitive attribute/property class. For UIA, we first filter out user identities that contain less than $2\times$ batch size number of samples and then split the dataset according to user identities. We select $15$ and $30$ users on the CREMA-D dataset, and $150$ and $300$ users on the CelebA dataset as the training and public sets, respectively. We select more users on the CelebA dataset because the majority of users only have very few samples ($\leq16$).
We set $m=6$ for DIA, i.e., inferring over $6$ ratio bins ($\{0\}, (0,0.2], (0.2,0.4], ..., (0.8, 1]$), and $m=5$ for UIA, i.e., choosing from $5$ candidate users.
For AIA and PIA, we assume the adversary has access to a prior of the sensitive variable that is estimated from the population. For DIA and UIA, we assume the adversary holds an uninformed prior, and thus the baseline is simply random guessing.
The default batch sizes are $16$ for AIA and PIA, $128$ for DIA, and $8$ for UIA.
For AIA, PIA, and DIA, the total number of trials $T$ of each experiment is equal to the number of random draws of training batches (i.e., $5{,}000$); for UIA, $T$ is the number of random draws of candidate sets, which we set to be $1{,}000$.
We repeat each experiment with $5$ different random seeds and report the mean and standard deviation of the results.

\subsection{Evaluation of Inference Attacks}
We evaluate each type of inference attack with a small shadow dataset ($1{,}000$ samples) and compare the results of single-round attacks (where the adversary only observes a single round of gradients) to multi-round attacks (where the adversary gets continual observation of the gradients).
Due to space limits, we only include a snapshot of the results (one dataset per attack) in Figure~\ref{fig:sr_mr_summary} and provide the full results in Appendix Figure~\ref{fig:sr_mr}.

\textbf{Attribute Inference.}
We present the results of AIA in Figure~\ref{fig:sr_mr_aia}. We observe that the adversary is able to infer the sensitive attribute with high confidence using only $1{,}000$ shadow data samples. For instance, on the Adult dataset, the multi-round adversary is able to achieve a high average AUROC of $0.9991$ and a TPR@$1\%$FPR of $0.9823$. On the Health dataset, however, the AUROC of the multi-round adversary reduces slightly to $0.8122$ while the TPR@$1\%$FPR drops drastically to $0.1611$. This is likely because the sensitive attribute on the Health dataset contains an ``unknown'' class ($18.9\%$) that is uncorrelated with other features, making it hard to estimate statistically.

\begin{figure*}[t]
    \centering
    \includegraphics[width=0.94\textwidth]{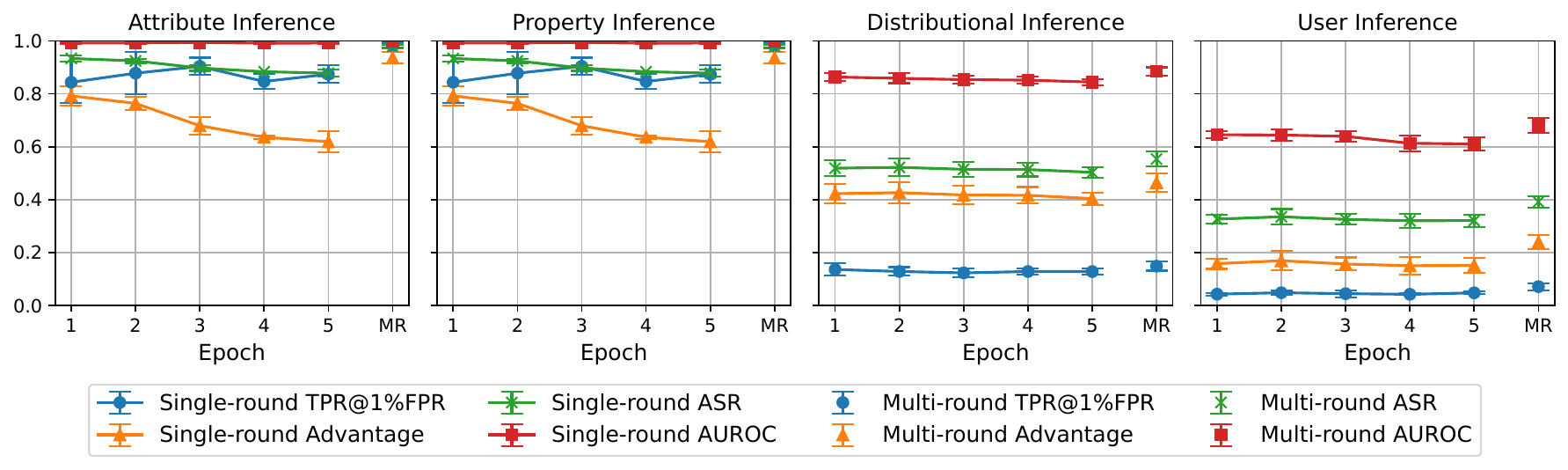}
    \caption{Comparison of single-round and multi-round inference attacks on the Adult (AIA, PIA, DIA) and CREMA-D (UIA) datasets. A complete result on all datasets is provided in Appendix Figure~\ref{fig:sr_mr}.}
    \label{fig:sr_mr_summary}
\end{figure*}

\textbf{Property Inference.}
Figure~\ref{fig:sr_mr_pia} depicts the results of PIA, where we observe that the adversary is able to achieve high performance across all five datasets.
Namely, the average AUROCs of the multi-round adversary on the Adult, Health, CREMA-D, CelebA, and UTKFace datasets are $0.9919$, $0.8294$, $0.8970$, $0.9993$, and $0.9167$, respectively.
This consistent high attack performance is in contrast to the general low correlation between the sensitive properties and the task labels across all datasets as indicated in Table~\ref{tab:data_summary} (except for CelebA, where a spurious relationship exists), which suggests that the information leakage observed is intrinsic to the computed gradients~\cite{melis2019exploiting}, regardless of the specific data type and learning task.

\textbf{Distributional Inference.}
Figure~\ref{fig:sr_mr_dia} summarizes the results of DIA.
Although distributional inference is a more challenging task ($6$-class ordinal classification), we observe that the multi-round adversary still performs fairly well with a batch size of $128$, achieving an average AUROC of $0.8848$, $0.7806$, $0.7572$, $0.9522$, and $0.7664$ on the Adult, Health, CREMA-D, CelebA, and UTKFace datasets, respectively.

\textbf{User Inference.}
We report the results of UIA in Figure~\ref{fig:sr_mr_uia}. We observe that the adversary is able to identify the user with relatively high confidence on the CelebA dataset, with an average AUROC and TPR@$1\%$FPR of $0.8935$ and $0.2828$ for the multi-round adversary. On the CREMA-D dataset, the average AUROC of the multi-round adversary is only $0.6808$, which may be due to the low identifiability of the features extracted for emotion recognition.

\textbf{General Observations.}
Additionally, we have the following general observations across different type of attacks and datasets.
First, the performance of single-round attacks decreases as the training progresses. This is because the gradients of the training data will become smaller in magnitude as the training loss decreases and thus the variation within these gradients will become harder to capture.
Second, on most datasets, the multi-round attack performs better than any single-round attack, proving the effectiveness of the Bayesian attack framework.
Third, we observe very similar performance for AIA and PIA on the tabular datasets. This indicates that whether the sensitive variable is internal or external to the data features does not affect the inference performance.

\subsection{Attack Analyses}
We investigate the following factors that may affect the performance of inference attacks.

\begin{figure*}[t]
    \centering
    \includegraphics[width=\linewidth]{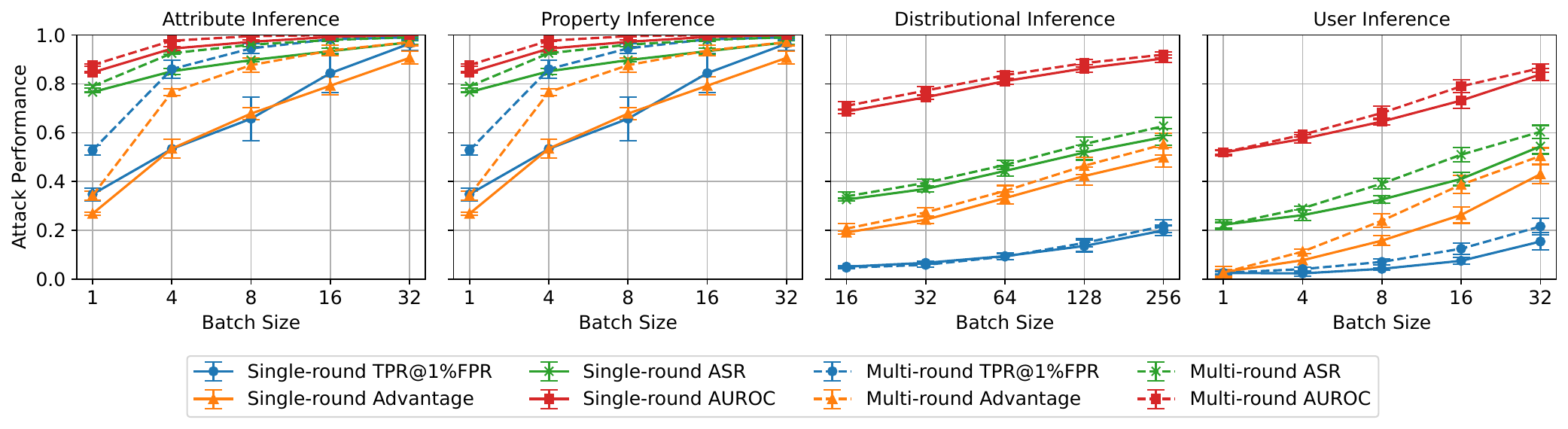}
    \caption{Sensitivity analysis of the impact of varying batch sizes on the performance of inference attacks.}
    \label{fig:batch_size}
\end{figure*}

\textbf{Impact of Batch Sizes.}
In Figure~\ref{fig:batch_size}, we study the impact of varying batch sizes on the performance of the inference attacks.
We report the results on the Adult dataset for AIA, PIA, and DIA, and results on the CREMA-D dataset for UIA.
\revise{We observe that the performance of all four considered inference attacks improves as the batch size increases.} This is because the records within the batch are sampled from the same conditional distribution $\prob(\rvx, \rvy|\rva)$.
\revise{As the private information $\va$ is shared across the batch, a larger batch size would amplify the private information and suppress other varying signals, thereby improving inference performance on $\va$.}
For distributional inference, the difference in the number of samples with the property between each ratio bin $\lfloor \alpha k \rfloor$ also increases as the batch size increases and thus becomes easier to distinguish.
For AIA and PIA, we observe that the gap between the single-round adversary (solid lines) and multi-round adversary (dashed lines) is the largest when the batch size is $4$, and then gradually reduces as the batch size increases further due to performance saturation.
\revise{This result suggests that simply aggregating more data does not protect gradients from inference. In fact, it may even increase the privacy risk in distributed learning where data are sampled from the same conditional distribution. This indicates that data aggregation alone is insufficient to achieve meaningful privacy in these settings.}

\begin{figure*}[t]
    \centering
    \includegraphics[width=\textwidth]{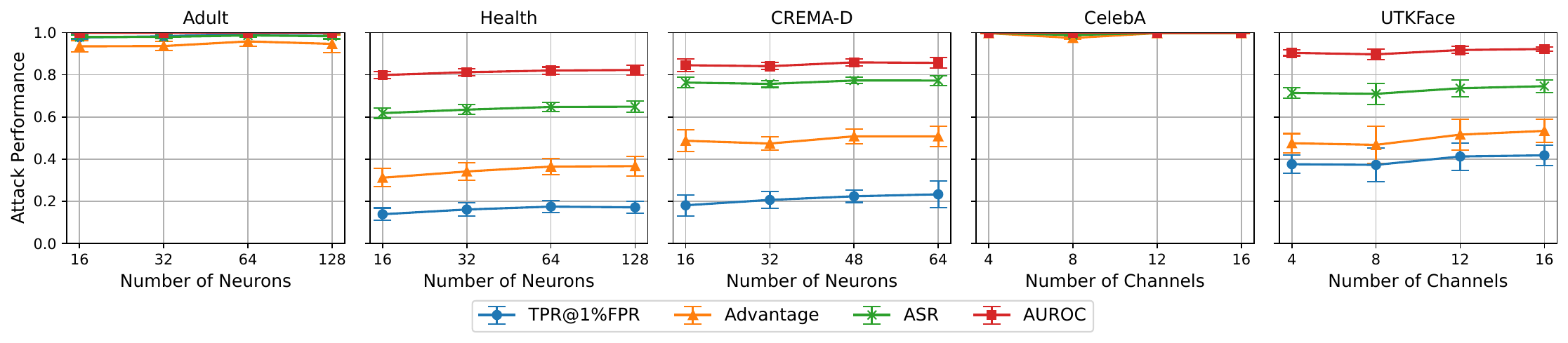}
    \caption{\revise{Sensitivity analysis of the impact of varying model sizes on the performance of \textit{Property Inference Attack}.}}
    \label{fig:model_size_PIA}
\end{figure*}

\begin{figure}[t]
    \centering
    \includegraphics[width=\linewidth]{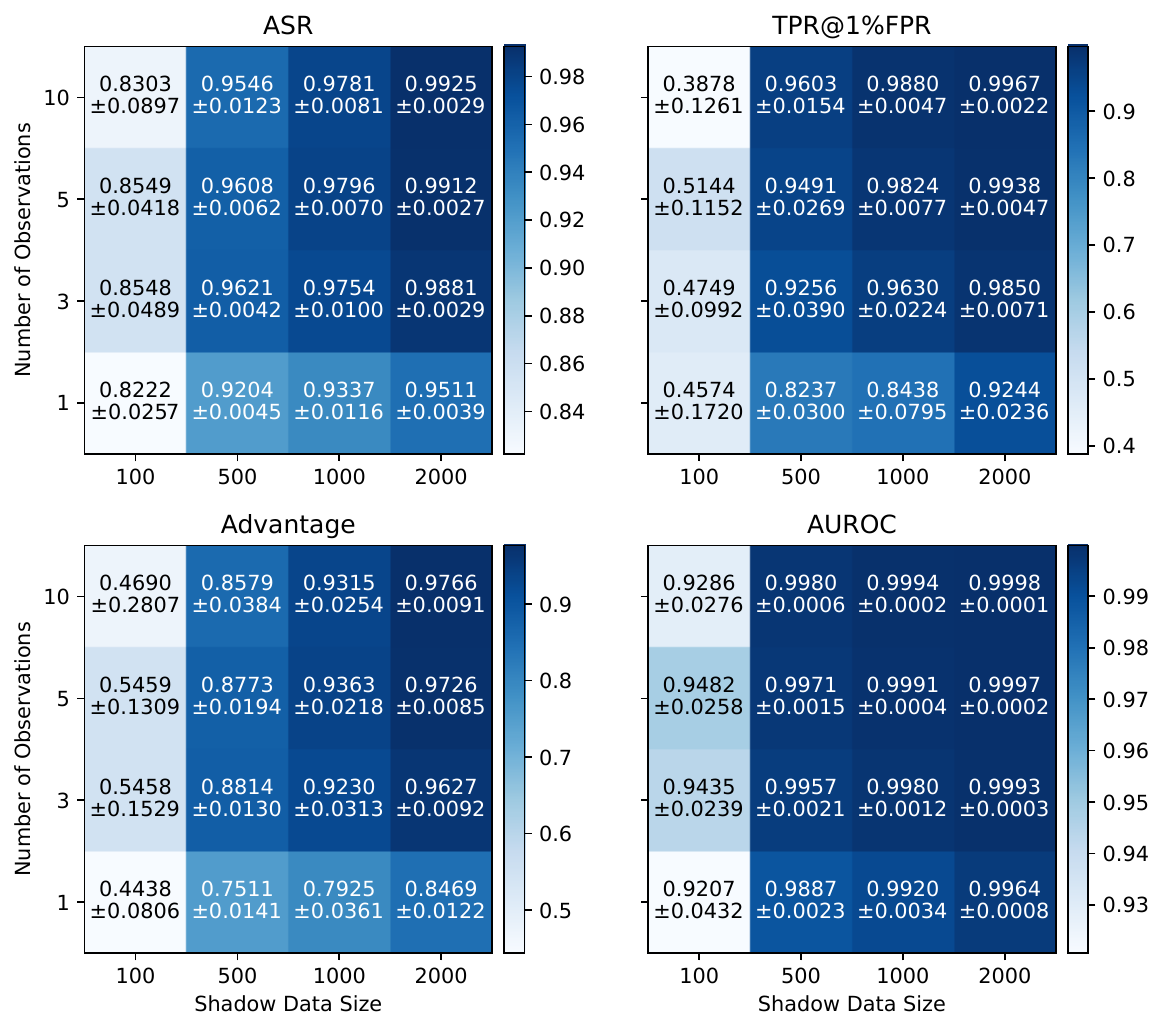}
    \caption{Sensitivity analysis of the impact of adversary's knowledge on the performance of \textit{Property Inference Attack} on the Adult dataset with a batch size of 16.}
    \label{fig:adv_knowledge}
\end{figure}

\textbf{Impact of Adversary's Knowledge.}
To investigate the impact of the adversary's knowledge on the performance of the attack, we use PIA as an example and plot the attack performance with varying shadow data size and number of observations on the Adult dataset in Figure~\ref{fig:adv_knowledge}.
We observe the general trend that the attack performance increases with the number of observations and available shadow data samples. Interestingly, the attack performance does not always increase monotonically along each axis. For instance, given a small shadow dataset of only $100$ samples, the AUROC of an adversary that observes $10$ rounds does not outperform an adversary that only observes $5$ rounds of gradients. This is likely because when the model is near convergence, the gradients are small and thus have low variance, which requires more shadow data to accurately estimate the posterior. Such errors in the predictive model will accumulate when using the summation of the log-likelihoods of all single rounds to approximate the joint distribution (Eq.~(\ref{eq:indepedence})), eventually leading to suboptimal performance.

\revise{\textbf{Impact of Model Size.}
In Figure~\ref{fig:model_size_PIA}, we use PIA as an example to study the impact of the machine learning model size. We control the size of the models by varying the model width. Specifically, for fully connected neural networks, we control the number of neurons for the hidden layer. For convolutional neural networks, we control the number of output channels for the first convolutional layer, with the remaining convolutional layers being scaled accordingly. 
We observe that the attack performance tends to improve slightly with increasing model size, except for the Adult and UTKFace datasets, where performance is saturated. However, most of these improvements are not statistically significant (falling within the margin of error) and thus do not allow for a conclusive statement.
We include additional results of other types of inference attacks in Appendix Figure~\ref{fig:model_size}, where we make similar observations. These results demonstrate that all four types of inference attacks can be generalized to larger model sizes.}

\section{Defenses}\label{sec:defense}

\revise{In this section, we investigate five types of strategies for defending inference from gradients against both static and adaptive adversaries and analyze their performance from an information-theoretic view. The main takeaways from our analyses are: (1) heuristic defenses can defend static adversaries but are ineffective against adaptive adversaries, (2) DP-SGD~\cite{abadi2016deep} is the only considered defense that remains effective against adaptive attacks, at the cost of sacrificing model utility, and (3) reducing the mutual information between the released gradients and the sensitive variable is a key ingredient for a successful defense.}

\subsection{Privacy Defenses Against Inference}

Privacy-enhancing strategies in machine learning generally follow two principles: \textit{data minimization} and \textit{data anonymization}.
\revise{Data minimization strategies, such as the application of cryptographic techniques (e.g., Secure Multi-party Computation and Homomorphic Encryption) and Federated Learning, aim to reveal only the minimal amount of information that is necessary for achieving a specific computational task - and only to the necessary parties. As shown by prior work~\cite{truex2019hybrid,elkordy2023much,lam2021gradient,kerkouche2023client}, data minimization alone may not provide sufficient privacy protection and, thus, should be applied in combination with data anonymization defenses to further reduce privacy risks.}
However, for heuristic-based privacy defenses, it is important to conduct a careful evaluation of their effectiveness against adaptive adversaries.
We consider the following five types of representative defenses from the current literature in our experiments:

\begin{enumerate}[leftmargin=*]

\item \textbf{Gradient Pruning.} Gradient pruning creates a sparse gradient vector by pruning gradient elements with small magnitudes. This strategy has been used as a baseline for privacy defense in federated learning~\cite{zhu2019deep,sun2021soteria,wu2023learning}. By default, we set the pruning rate to be $99\%$.

\item \textbf{SignSGD.} SignSGD~\cite{bernstein2018signsgd} binarizes the gradients by applying an element-wise sign function to the gradients,
thereby compressing the gradients to 1-bit per dimension. Similar to gradient pruning, it has been explored in prior work~\cite{wu2023learning,yue2023gradient} as a defense against data reconstruction attacks in federated learning.
\revise{
Along similar lines, Kerkouche \textit{et al.}~\cite{kerkouche2020federated} evaluated SignFed, a variant of the SignSGD protocol adapted for federated settings, and found it to be more resilient to privacy and security attacks than the standard federated learning scheme.}

\item \textbf{Adversarial Perturbation.}
Inspired by prior research on protecting privacy through adopting evasion attacks in adversarial machine learning~\cite{jia2018attriguard,jia2019memguard,shan2020fawkes,o2022voiceblock}, we explore a heuristic defense strategy against inference attacks that inject adversarial perturbation to the gradients. Specifically, at each round of observation, the adversary first trains a neural network $f_\vphi: \rvg \rightarrow \rva$ to classify the sensitive variable $\va$ from the gradient $\vg$ using a public dataset (same as the shadow dataset). Then, the defense generates a protective adversarial perturbation to cause $f_\vphi$ to misclassify the perturbed gradients. We adopt $l_\infty$-bounded projected gradient descent (PGD)~\cite{madry2018towards}, which generates the adversarial example $\vg'$ (perturbed gradient) by iteratively taking gradient steps.
For AIA, PIA, and DIA, this defense generates an untargeted adversarial perturbation through gradient ascent, i.e.,
$\Tilde{\vg}\leftarrow\prod_{\gB_\infty(\vg, \gamma)}\big(\Tilde{\vg} + \alpha \cdot \text{sign}( \nabla_{\vg}\mathcal{L}(\vphi, \vg, \va))\big)$, where $\gB_\infty(\vg, \epsilon)$ is the $l_\infty$ norm ball centered around $\vg$ with radius $\epsilon$.
For UIA, the defense generates a targeted adversarial perturbation through gradient descent, i.e.,
$\Tilde{\vg}\leftarrow\prod_{\gB_\infty(\vg, \gamma)}\big(\Tilde{\vg} - \alpha \cdot \text{sign}( \nabla_{\vg}\mathcal{L}(\vphi, \vg, \va_t))\big)$, to make the gradients misrecognized as the target user $\va_t$.
By default, we set the total number of steps to be $5$, $\gamma=0.005$, and $\alpha=0.002$.

\item \textbf{Variational Information Bottleneck (VIB).} This defense inserts an additional VIB layer~\cite{alemi2016deep} that splits the neural network $f_\vtheta$ into a probabilistic encoder $p(\rvh|\rvx)$ and a decoder $q(\rvy|\rvh)$, where $\rvh$ is a latent representation that follows a Gaussian distribution.
An additional Kullback-Leibler (KL) divergence term is introduced to the training loss: $\mathcal{L}_{VIB}=\mathcal{L}(\vtheta, \gD) + \beta \cdot KL(p(\rvh|\rvx)||q(\rvz))$, where $q(\rvz)=\mathcal{N}(\bm{0}, \bm{I})$ is the standard Gaussian. Optimizing this VIB objective reduces the mutual information $I(\rvx;\rvh)$ between the representation and the input by minimizing a variational upper bound. Prior work suggests that this helps to reduce the model's dependence on input's sensitive attributes and improve privacy~\cite{song2019overlearning,scheliga2022precode,scheliga2023privacy}. We set $\beta=0.01$ as the default for our experiments.

\item \textbf{Differential Privacy (DP-SGD).}
Differential privacy (DP)~\cite{dwork2006calibrating} provides a rigorous notion of algorithmic privacy.
\definition{\textit{\textbf{($\varepsilon,\delta$)-Differential Privacy.}}}\label{def:dp}
An algorithm $\gM$ is said to satisfy ($\varepsilon,\delta$)-DP if for all sets of events $S$ defined on the output of $\gM$ 
and all neighboring datasets $\gD,\gD'$ that differ in one sample, the following inequality holds:
\begin{equation*}
    \prob(\gM(\gD) \in S) \leq e^\varepsilon \prob(\gM(\gD') \in S) + \delta.
\end{equation*}

\noindent The most widely adopted DP algorithm for training ML model is DP-SGD~\cite{abadi2016deep}.
At each step of training, the DP-SGD algorithm first clips the $l_2$ norm of per-sample gradients $\Tilde{\vg}_i \leftarrow \vg_i/\max(1, \frac{||\vg_i||_2}{\Delta})$
and then injects calibrated Gaussian noise to get the aggregated gradients $\Tilde{\vg} \leftarrow \frac{1}{k} \sum_{i=0}^k \big(\Tilde{\vg}_i + \gN(\bm{0}, \sigma^2\bm{I})\big)$.
DP-SGD achieves ($\varepsilon,\delta$)-differential privacy for any $\delta > 0$ with $\varepsilon=\Delta\sqrt{2\log\frac{1.25}{\delta}}/\sigma$ for each step, while the total privacy loss can be obtained through composition.
By default, we set $\Delta=2$ and $\sigma=0.1$ \revise{(corresponds to per-step $\varepsilon=96.90$ when $\delta=10^{-5}$)}.
\end{enumerate}

\begin{figure*}[t]
    \centering
    \includegraphics[width=\textwidth]{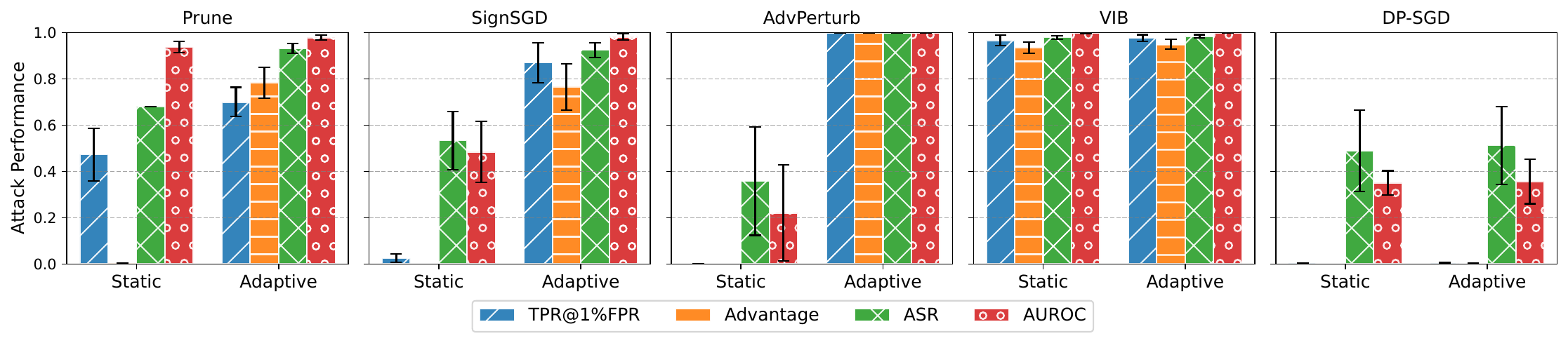}
    \caption{Comparison of various defenses against static and adaptive \textit{Property Inference Attack} on the Adult dataset with a batch size of 16. A complete result of all inference attacks is provided in Appendix Figure~\ref{fig:defenses_full}.}
    \label{fig:defenses_short_version}
\end{figure*}

\subsection{Defense Evaluation}

In Figure~\ref{fig:defenses_short_version}, we compare the performance of defenses against static and adaptive adversaries. Due to space limits, here we focus on PIA on the adult dataset. The full results including all four types of inference attacks are available in Appendix Figure~\ref{fig:defenses_full}.
We observe that heuristic defenses such as Gradient Pruning, SignSGD, and Adversarial Perturbation can successfully defend against static adversaries in terms of reducing the advantage of the adversary to zero. However, these defenses are ineffective against adaptive adversaries aware of the defense. For instance, in the case of gradient pruning, the adaptive adversary can achieve a high advantage ($0.7841$) that is only slightly decreased compared to no defense ($0.9363$).
Interestingly, in the case of Adversarial Perturbation, we found that the adaptive adversary's performance is increased, rather than decreased, compared to no defense, reaching a perfect advantage and AUROC of $1.00$.
For the rest of the defenses, namely, VIB and DP-SGD, the attack performance is consistent across static and adaptive adversaries. However, only DP-SGD manages to effectively reduce the advantage of the adaptive adversary to near zero.

To understand the privacy-utility trade-off of these defenses, we plot the PIA adversary's advantage evaluated on the training data versus the measured AUROC of the network on predicting the task label on the test dataset on the Adult dataset in Figure~\ref{fig:tradeoff}. We consider three different sets of parameters for each type of defense (details in Appendix). We observe that in the case of static adversaries, SignSGD achieves the best trade-off that approximates the ideal defense (upper left corner) by reducing the advantage to zero without affecting model utility. However, in the case of adaptive adversary, only DP-SGD provides a meaningful notion of privacy, at the cost of diminishing model utility.
Moreover, there may exist stronger adversaries that are more resilient against these defenses.
For instance, in Table~\ref{tab:pca_against_dp}, we show that an adversary using principal component analysis (PCA) with $50$ principal dimensions as dimensionality reduction can bypass the DP-SGD defense with \revise{$\varepsilon=96.90$ and $\delta=10^{-5}$} that defends an adversary using max-pooling, and requires $15\times$ larger noise to thwart.

In the next section, we analyze the underlying principles of these defenses and the necessary ingredients for a successful defense.

\begin{figure}[t]
    \centering
    \includegraphics[width=\linewidth]{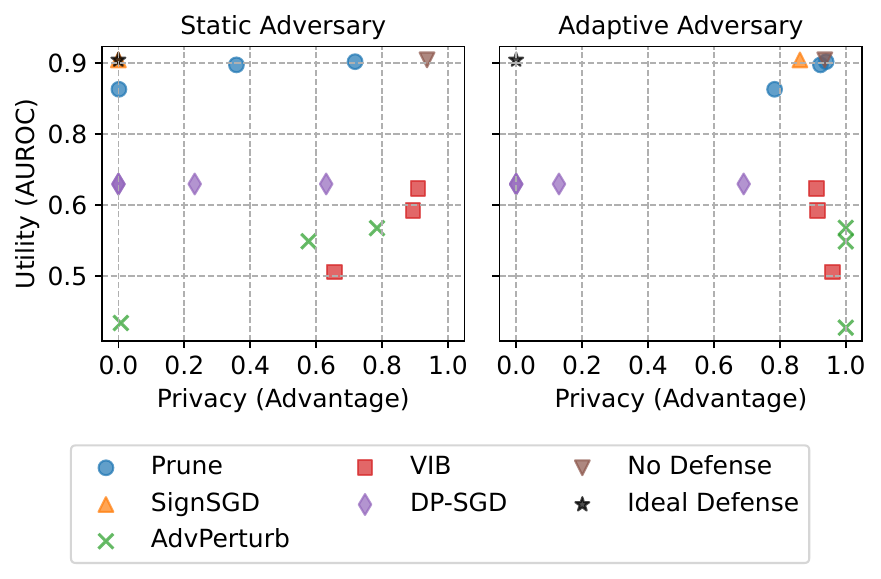}
    \caption{Privacy-Utility trade-off of various defenses against \textit{Property Inference Attack} on the Adult dataset.}
    \label{fig:tradeoff}
\end{figure}

\subsection{Defense Analyses}
In this section, we provide an information-theoretic perspective for understanding and analyzing defenses against inference attacks from gradients.

\textbf{Information-theoretic View on Inference Privacy.}
The inference attacks captured in the unified game can be viewed as performing statistical inference~\cite{du2012privacy} on properties of the underlying data distributions upon observing samples of the gradients.
A well-known information-theoretic result for analyzing inference is Fano's inequality, which guarantees a lower bound on the estimation error of any inference adversary.
Formally, consider any arbitrary data release mechanism that provides $\rmY$ computed from the private discrete random variable $\rmX$ supported on $\gX$.
Any inference from the observation $\rmY$ must produce an estimate $\hat{\rmX}$ that satisfies the Markov chain $\rmX \rightarrow \rmY \rightarrow \hat{\rmX}$. 
Let $\rve$ be a binary random variable that indicates an error, i.e., $\rve=1$ if $\hat{\rmX} \neq \rmX$. Then we have
\begin{equation} \label{eq:tight-fano}
    H(\rmX|\rmY) \leq H(\rmX|\hat{\rmX}) \leq H_2(\rve) + \prob(\rve=1) \log(|\gX|-1),
\end{equation}
where $H_2(\rve)=-\prob(e=1)\log\prob(e=1) - \big(1-\prob(e=1)\big)\log\big(1-\prob(e=1)\big)$ is the binary entropy.
For $|\gX|>2$, a standard treatment is to consider the mutual information $I(\rmX; \rmY) = H(\rmX) - H(\rmX|\rmY)$ and $H_2(\rve) \leq \log 2$, and thereby we can obtain a lower bound on the error probability:
\begin{equation} \label{eq:error}
    \prob(\hat{\rmX} \neq \rmX) \geq \frac{H(\rmX)-I(\rmX; \rmY) - \log 2}{\log (|\gX|-1)}.
\end{equation}
Note that this bound is vacuous when $|\gX|=2$, and a slightly tighter bound can be obtained by considering $H_2(\rve)$ exactly (rather than using the approximating bound of $\log 2$) and numerically computing the lowest error probability that satisfies the inequality in~(\ref{eq:tight-fano}), as noted by prior work~\cite{guo2023analyzing}.
The bound in inequality (\ref{eq:error}) captures both the prior (via $H(\rmX)$) and the cardinality of the sensitive variable alphabet, indicating that data with a large degree of uncertainty is hard to infer or reconstruct, which aligns with intuition from Balle \textit{et al.}~\cite{balle2022reconstructing}.
Inequality (\ref{eq:error}) generically holds for any data release mechanism. In the context of inference from gradients, the adversary's goal is to obtain an estimate of $\rva$ upon observing $\Tilde{\rvg}$, which can be described as a Markov chain of $\rva \rightarrow \rvx \rightarrow \rvg \rightarrow \Tilde{\rvg} \rightarrow \hat{\rva}$.
Since the adversary's success rate is $p=1-\prob(\rve=1)$, one can get an immediate upper bound on the adversary's advantage:
\begin{equation}
    \texttt{Adv}(p) \leq 1 - \frac{H(\rva) - I(\rva; \Tilde{\rvg}) - \log 2}{(1-p^*)\log (m-1)}.
\end{equation}
As $H(\rva)$ is a constant, this indicates that reducing $I(\rva; \Tilde{\rvg})$ results in increasing the lower bound of the error probability and consequently diminishing the adversary's advantage.
\revise{This analysis can be generalized to continuous sensitive variables by applying continuum Fano’s inequality~\cite{duchi2013distance}.}

\textbf{Understanding Defenses.}
Next, we provide an explanation of the failures of heuristic defenses using the above framework and argue that a successful defense should effectively minimize the mutual information $I(\rva; \Tilde{\rvg})$ between the gradients and the sensitive variable.
The Gradient Pruning and SignSGD defenses can be viewed as trying to reduce the number of transmitted bits in the gradients. However, this does not necessarily reduce the mutual information.
The neural network classifier $f_\vphi: \rvg \rightarrow \rva$ used in the Adversarial Perturbation defense is trained to minimize cross-entropy loss,
which provides an approximate upper bound on the conditional entropy $H(\rva | \rvg)$, and serves as a proxy for estimating the mutual information $I(\rva; \Tilde{\rvg}) = H(\rva) - H(\rva | \rvg)$.
However, generating adversarial perturbations to produce $\Tilde{\rvg}$ against this fixed classifier does not necessarily result in a reduction of the mutual information $I(\rva; \Tilde{\rvg})$, and likely increases it.
This is because the gradient steps $\nabla_{\vg}\mathcal{L}(\vphi, \vg, \va)$ used to generate the protective perturbation also contain information about $\va$. As the perturbation generation process is deterministic, an adaptive adversary can learn to pick up these patterns and gain additional advantage.
\revise{In the case of VIB, the mechanism is stochastic but optimizing the VIB objective only gradually reduces the mutual information $I(\rvx;\rvh)$ between the latent representation $\rvh$ and the input $\rvx$, which still does not guarantee a reduction in $I(\rva;\Tilde{\rvg})$ during the optimization process.}
By design, differential privacy is not intended to protect against statistical inference as its goal is to preserve the statistical properties of the dataset while protecting the privacy of individual samples.
However, an alternative information-theoretical interpretation of differential privacy is that it places a constraint on mutual information~\cite{bun2016concentrated,cuff2016differential}. An easy way to see this is that by adding Gaussian noises to the gradients, the DP-SGD algorithm essentially creates a Gaussian channel between the true and released gradients, thereby placing a constraint on $I(\rvg; \Tilde{\rvg})$, which further bounds $I(\rva; \Tilde{\rvg})$ as $I(\rva; \Tilde{\rvg}) \leq I(\rvg; \Tilde{\rvg})$ according to the data processing inequality.
More concretely, due to
the Gaussian channel $\Tilde{\rvg} = \rvg + \gN(\bm{0}, \sigma^2\bm{I})$,
we have the upper bound given by the channel capacity
$I(\rvg;\Tilde{\rvg}) \leq \frac{1}{2}\log(1 + \frac{P}{\sigma})$, if the gradients $\vg$ satisfy an average power constraint $\E[ \| \vg \|_2^2 ] \leq nP$, where $n$ is the dimensionality of $\vg$.
One can obtain a stronger result in cases where the $l_2$ sensitivity is bounded (e.g., Theorem 2 in \cite{guo2023analyzing}).

\revise{It is worth noting that the goal of our analyses here is to provide a perspective for understanding the effectiveness of a class of defense strategies, rather than deriving tight bounds.} 
Additionally, as mutual information is a statistical quantity, the mutual information interpretation of inference privacy inherently only captures the average-case privacy risk.
In the next section, we provide a privacy auditing framework for empirically estimating the privacy risk by approximating the worst-case scenario.

\begin{table}[t]
\centering
\caption{Comparison of different dimensionality reduction strategies in PIA on Adult against DP-SGD defense \revise{($\delta=10^{-5}$)}.}
\label{tab:pca_against_dp}
\resizebox{\linewidth}{!}{
\begin{tabular}{c|c|cccc}
\toprule
\textbf{\revise{$\varepsilon$}} & \textbf{Adversary Type} & \textbf{AUROC}                                            & \textbf{TPR@1\%FPR}                                       & \textbf{ASR}                                         & \textbf{Advantage}                                        \\ \toprule
\revise{96.90}            & MaxPooling              & \begin{tabular}[c]{@{}c@{}}0.3004\\$\pm$0.0773\end{tabular} & \begin{tabular}[c]{@{}c@{}}0.0017\\$\pm$0.0010\end{tabular} & \begin{tabular}[c]{@{}c@{}}0.5732\\$\pm$0.1124\end{tabular} & \begin{tabular}[c]{@{}c@{}}0.0001\\$\pm$0.0002\end{tabular} \\ \midrule
\revise{96.90}            & PCA                     & \begin{tabular}[c]{@{}c@{}}0.9825\\$\pm$0.0112\end{tabular} & \begin{tabular}[c]{@{}c@{}}0.7284\\$\pm$0.1679\end{tabular} & \begin{tabular}[c]{@{}c@{}}0.9437\\$\pm$0.0222\end{tabular} & \begin{tabular}[c]{@{}c@{}}0.8239\\$\pm$0.0694\end{tabular} \\ \midrule
\revise{6.46}              & PCA                     & \begin{tabular}[c]{@{}c@{}}0.7010\\$\pm$0.0278\end{tabular} & \begin{tabular}[c]{@{}c@{}}0.0471\\$\pm$0.0120\end{tabular} & \begin{tabular}[c]{@{}c@{}}0.6995\\$\pm$0.0091\end{tabular} & \begin{tabular}[c]{@{}c@{}}0.0598\\$\pm$0.0286\end{tabular}\\ \bottomrule
\end{tabular}
}
\end{table}

\section{Empirical Estimation of Privacy Risk}
\label{sec:privacy_audit}

In the privacy game defined in Definition~\ref{def:unified_game}, the data is randomly sampled from the distribution, which only captures the average-case privacy risk and therefore cannot be used for reasoning about the minimal level of noise required for ensuring a certain level of privacy, as it may underestimate the privacy risk in the worst case.
To better understand the privacy risk in the worst-case scenario, we provide a privacy auditing framework for empirically estimating the privacy leakage of a specific type of inference attack, namely, attribute inference, by allowing the data to be chosen adversarially.
We start with a formal definition of per-attribute privacy following prior work~\cite{ahmed2016social,ghazi2022algorithms}:

\begin{definition}{\textit{\textbf{Per-attribute DP.}}}
A randomized mechanism $\gM$ is \revise{$(\varepsilon, \delta)$-per-attribute DP} if for all pairs of inputs $x,x'$ differing only on a single attribute and for all events $S$ defined on the output of $\gM$, the following inequality holds:
\begin{equation*}
    \prob[\gM(x)\in S] \leq e^\varepsilon \cdot \prob[\gM(x')\in S] + \delta.
\end{equation*}
\end{definition}

One can show that DP-SGD satisfies \revise{$(\varepsilon, \delta)$-per-attribute DP}. However, it is hard to derive the privacy parameter analytically, as the per-attribute sensitivity of the gradient is not readily tractable and the common technique of gradient clipping only provides a very loose bound on sensitivity.
Instead, we seek to obtain an empirical estimate of the per-attribute DP for \textit{each step} through the following audit game.

\definition{\textit{\textbf{Per-Attribute Privacy Audit Game.}}} Suppose $\rva \in [m]$ is a discrete attribute that takes on $m$ values. The per-attribute privacy audit game between a challenger (private learner) and an adversary (auditor) is as follows:

\begin{enumerate}[label={(\arabic*)}]
    \item Adversary chooses a record $\vz$ with attribute value $\va$.

    \item Challenger samples a uniformly random private bit $\rb \in \{0, 1\}$. If $\rb = 1$, assign the attribute in $\vz$ with a new value uniformly sampled from $[m]\backslash\{\va\}$.

    \item Challenger obtains the latest model parameters as $\vtheta$ through the training algorithm $\gT$.

    \item Challenger computes the gradient of the record, $\vg = \nabla_{\vtheta} \mathcal{L}(\vtheta, \vz)$.
    
    \item Challenger applies the DP-SGD algorithm $\mathcal{M}$ to produce a privatized version of the gradient $\Tilde{\vg} = \mathcal{M}(\vg)$.

    \item The adversary $\gA$ gets access to $\gL$, $\gT$, and the model parameters $\vtheta$, released gradients $\vg$, the auxiliary information about the record $\varphi(\vz)$, and the defense mechanism $\mathcal{M}$.

    \item The adversary outputs the inferred information $\hat{\rb}$, i.e., $\hat{\rb}\leftarrow \gA(\gL, \gT, \vtheta, \vg, \varphi(\rvz), \mathcal{M}).$ The adversary wins if $\hat{\rb} = \rb$ and loses otherwise.
\end{enumerate}

There are two major differences between the audit game and the inference game as defined in Definition~\ref{def:unified_game}.
First, the record is chosen by the adversary, instead of being randomly drawn from the distribution, which aims to simulate the worst-case scenario over all adjacent input pairs as captured by the per-attribute DP definition.
Second, instead of having access to distributional information $\prob(\rvx, \rvy, \rva)$, the adversary gets access to some auxiliary information about the record $\varphi(\vz)$, which is assumed to be all the remaining features except for $\va$. This is to approximate the strong adversarial assumption in per-attribute DP where the adversary has access to everything except for one attribute.

\textbf{Empirical Privacy Estimate.}
Analogous to the operational interpretation of canonical DP~\cite{kairouz2015composition}, we can interpret attribute DP as a hypothesis test with $\rb =0$ as the null hypothesis ($\gH_0$) and $\rb=1$ as the alternative hypothesis ($\gH_1$).
We compute the test statistics $\rt(\Tilde{\vg})=||\Tilde{\vg} - \vg_{\gH_0}||_2$ using the $l_2$ norm between the observed gradient $\Tilde{\vg}$ and the hypothetical gradient $\vg_{\gH_0}$ under $\gH_0$ (i.e., $\vg_{\gH_0} = \vg/\max(1, \frac{||\vg||_2}{\Delta})$ when $\varphi(\vz)=\va$). This is connected to the likelihood of observing $\Tilde{\vg}$ under $\gH_0$ since $\Tilde{\vg}_{\gH_0} \sim \gN(\vg_{\gH_0}, \sigma^2\bm{I})$.
We then execute the game several times to get an empirical distribution of the test statistics.
Building on prior works on auditing canonical DP with membership inference attacks~\cite{jagielski2020auditing,nasr2021adversary,maddock2022canife,andrew2023one}, finally, we derive the \revise{empirical privacy loss parameter $\hat{\varepsilon}$} given the false positive rate (FPR) and false negative rate (FNR) at critical value $c$ as
\begin{equation*}
    \hat{\varepsilon} = \max \bigg( \log \frac{1-\delta-\text{FPR}}{\text{FNR}}, \log \frac{1-\delta-\text{FNR}}{\text{FPR}} \bigg),
\end{equation*}
where the critical value $c$ is chosen over all possible values to maximize the empirical estimate $\hat{\varepsilon}$ to obtain a worst-case measure. Similar to previous work~\cite{nasr2021adversary}, we additionally compute and report the $95\%$ confidence intervals for $\hat{\varepsilon}$ using the Clopper-Pearson method.

\textbf{Crafting the Worst-case Sample.}
To further improve the estimate, we approximate the worst-case scenario by crafting a canary record $\vz^*$ to maximize the expected difference in the test statistics between $\gH_0$ and $\gH_1$, via the optimization 
\begin{equation*}
    \vz^* = \argmax_\vz \text{Dist}(\vg_{\gH_0}, \vg_{\gH_1}),
\end{equation*}
where $\text{Dist}(\cdot, \cdot)$ is a distance measure. We experimented with cosine similarity and mean squared error (MSE) and found that MSE performs better empirically.

\begin{figure}[t]
     \centering
     \begin{subfigure}[b]{0.47\linewidth}
         \centering
         \includegraphics[width=\textwidth]{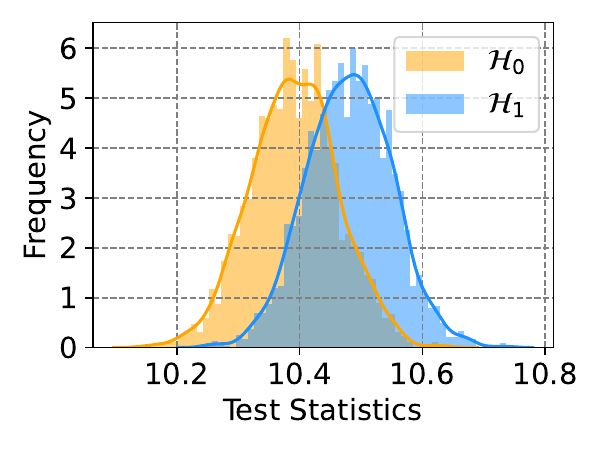}
         \caption{Randomly sampled record.}
     \end{subfigure}
     \hspace{2mm}
     \begin{subfigure}[b]{0.47\linewidth}
         \centering
         \includegraphics[width=\textwidth]{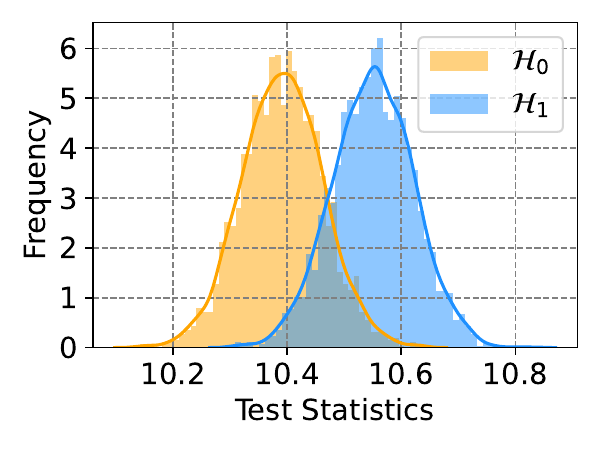}
         \caption{Adversarially crafted record.}
     \end{subfigure}
    \caption{Comparison of the test statistics distribution from auditing games with different choices of test record $\vz$.}
    \label{fig:test_statistics_distribution}
\end{figure}

\begin{figure}[t]
     \centering
     \includegraphics[width=\linewidth]{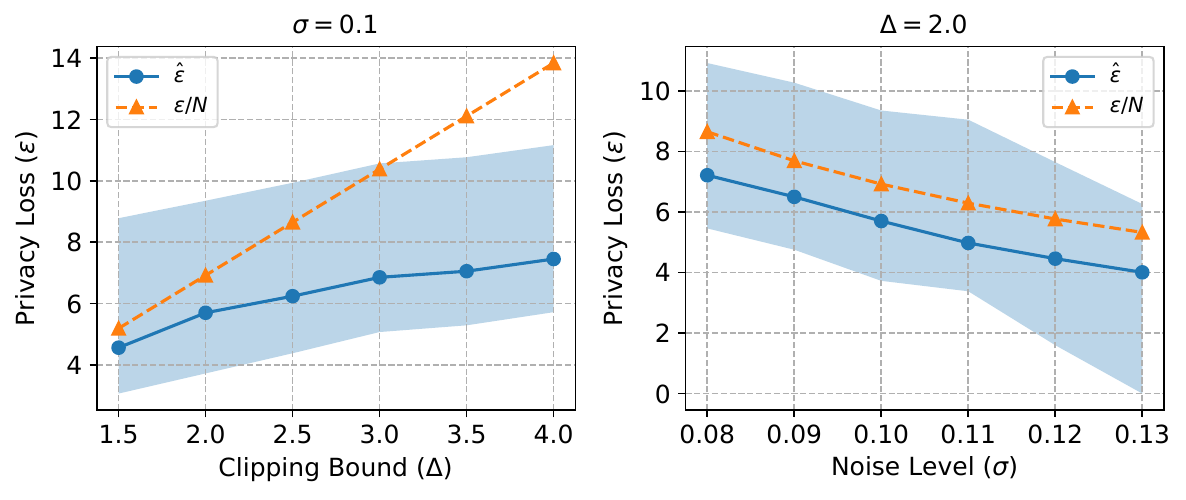}
    \caption{Comparison of the empirical per-attribute privacy loss ($\hat{\varepsilon}$) with $95\%$ confidence interval and the theoretical privacy loss ($\varepsilon$) normalized by the total number of attributes $N$.}
    \label{fig:emprirical_eps}
\end{figure}

\textbf{Empirical Results.}
We first conduct experiments on the Adult dataset using a fully-connected neural network with one hidden layer of $100$ neurons to verify the effectiveness of the adversarially crafted sample.
We compute the test statistics for $5{,}000$ trials with $\Delta=2$ and $\sigma=0.1$ and plot the histogram of the test statistics in Figure~\ref{fig:test_statistics_distribution}. We observe that the distributions of test statistics under $\gH_0$ and $\gH_1$ are more separable using the adversarially crafted canary record, compared to a randomly drawn record from the data distribution, thereby providing a better estimate on the worst-case privacy risk.
We then compare \revise{the empirical estimated $\hat{\varepsilon}$ to the theoretical $\varepsilon$} computed with the gradient clipping bound $\Delta$ as the per-attribute sensitivity, at $\delta=10^{-5}$, using adversarially crafted records.
Figure~\ref{fig:emprirical_eps} plots \revise{the empirically estimated $\hat{\varepsilon}$ and the theoretical $\varepsilon$} normalized by the total number of attributes ($N=14$) with varying clipping bound $\Delta$ and noise level $\sigma$.
We observe that using the clipping bound as the per-attribute sensitivity indeed leads to a very conservative estimate of the privacy loss, with a large gap \revise{(${\varepsilon}/{\hat{\varepsilon}}= 1.86 N$)} when $\sigma=0.1$ and $\Delta=4$. As the clipping bound reduces, the ratio gradually approximates to $N$ \revise{(${\varepsilon}/{\hat{\varepsilon}}=1.14 N$ when $\Delta=1.5$)}. When the clipping bound is fixed to $\Delta=2$, the gap is relatively consistent across different noise levels \revise{(e.g., ${\varepsilon}/{\hat{\varepsilon}}=1.20 N$ when $\sigma=0.08$ and ${\varepsilon}/{\hat{\varepsilon}}=1.33 N$ when $\sigma=0.13$)}.

\section{Conclusion and Discussion}
\label{sec:discussion}

We conduct a systematic analysis of private information leakage from gradients under different levels of adversarial uncertainties within a unified inference framework.
We provide an information-theoretic perspective for explaining and analyzing the efficacy of defenses for preventing inference through the gradient channel. Finally, we introduce an auditing approach for estimating realistic privacy risks against attribute inference.
\revise{There are three primary takeaways from this study: (1) data aggregation alone does not provide sufficient privacy in distributed learning, (2) reducing the mutual information is a key ingredient of successful defenses against inference from gradients, and (3) it is important to specify the privacy context (e.g., average vs. worst-case vulnerability of a dataset) when estimating privacy risks.}

Our findings open up several interesting discussions.
Firstly, information leakage from the gradient exhibits distinct characteristics compared to leakage from the model parameters, suggesting a competing relationship. In the extreme case, a perfectly memorized sample could pose high privacy risks via model parameters, yet disclose no information through gradients if the loss is zero.
Secondly, while the quantification of inference attack risks with mutual information provides theoretical guarantees and a guiding principle for privacy mechanism design, there are practical disadvantages in tractability and composition.
As mutual information is a statistical quantity that depends on the unknown data distribution, practical application requires estimation from data, which may be challenging in distributed learning scenarios.
Further, addressing information leakage across multiple rounds requires dealing with the mutual information between the private variable and all gradient observations handled jointly, i.e., $I(\rva ; \rvg_1, \ldots, \rvg_r)$, where $\rvg_i$ denotes the gradients observed in each round.
However, a complication of this combined mutual information is that it cannot simply be exactly decomposed as a summation of single-round mutual information terms $I(\rva ; \rvg_i)$, i.e., mutual information lacks a convenient composition property.
\revise{Finally, we demonstrated that among the defenses considered, only DP-SGD provides a meaningful notion of privacy against adaptive adversaries, despite affecting model utility. However, the loss in utility can softened by improving the algorithms. For instance, recent research~\cite{aerni2024evaluations} showed in the centralized setting that state-of-the-art DP-SGD solutions provide better privacy than most empirical defenses at similar utility.}
\revise{Improving multi-round privacy analysis and tuning defenses (e.g., by minimizing the mutual information objective) towards better privacy-utility trade-offs using public data are interesting avenues for future research.}

\begin{acks}
This research was sponsored in part by the National Institute of Health (NIH) under grant number U54HG012510.
\end{acks}

\bibliographystyle{ACM-Reference-Format}
\bibliography{reference}

\appendix
\pagebreak

\section{Detailed Experimental Setup}

\subsection{Details About ML Models}
For the Adult and Health datasets, we train a fully-connected neural network with two hidden layers of sizes $32$ and $16$. For the CREMA-D, we train a fully-connected neural network with two hidden layers of $64$ neurons. For the CelebA and UTKFace datasets, we train a convolutional neural network with $9$ convolutional layers and $2$ MaxPooling layers (details in Table~\ref{tab:cnn_arch}). ReLU is used as the default activation function for all models.

\begin{table}[h]
    \centering
    \caption{Model architecture for image datasets.}
    \vspace{-1mm}
    \label{tab:cnn_arch}
    \resizebox{0.6\linewidth}{!}{
    \begin{tabular}{cccc}
    \hline
    \textbf{Layer}   & \textbf{Kernel} & \textbf{Stride} & \textbf{Output} \\ \hline \hline
    Conv2D & $3\times3$    & $1\times1$    & $8$     \\
    Conv2D & $3\times3$    & $1\times1$    & $16$     \\
    Conv2D & $3\times3$    & $1\times1$    & $16$     \\
    Conv2D & $3\times3$    & $1\times1$    & $32$     \\
    Conv2D & $3\times3$    & $1\times1$    & $32$     \\
    Conv2D & $3\times3$    & $1\times1$    & $32$     \\
    MaxPool2D & $3\times3$    & $3\times3$    & $32$     \\
    Conv2D & $3\times3$    & $1\times1$    & $32$     \\
    Conv2D & $3\times3$    & $1\times1$    & $32$     \\
    Conv2D & $3\times3$    & $1\times1$    & $32$     \\
    MaxPool2D & $3\times3$    & $3\times3$    & $32$     \\
    Flatten     &    $-$    &    $-$    & $-$       \\
    FC     &    $-$    &    $-$    & \# of classes       \\
    \hline
    \end{tabular}
    }
    \vspace{-1mm}
\end{table}

\subsection{Parameter Choices for Privacy-Utility Analysis of Defenses}
For each type of defense, we consider three different sets of parameters in Figure~\ref{fig:tradeoff}: Gradient Pruning with $90\%$, $95\%$, and $99\%$ rate; Adversarial Perturbation with $(\gamma=5\times10^{-4},\alpha=2\times10^{-4})$, $(\gamma=1\times10^{-3},\alpha=3\times10^{-4})$, and $(\gamma=5\times10^{-3},\alpha=2\times10^{-3})$; VIB with $\beta=10^{-1},10^{-2},10^{-3}$; and DP-SGD with $(\sigma=1\times10^{-1},\Delta=2)$, $(\sigma=2\times10^{-2},\Delta=2)$, and $(\sigma=1\times10^{-2},\Delta=2)$.

\subsection{Details About Crafting The Canary Record}
We generate synthetic canary records by initializing $\rz$ using a vector sampled from the standard normal distribution and then solve the optimization using the Adam optimizer with a learning rate of $5\times10^{-2}$ for $2{,}000$ iterations.

\section{Additional Results}

\subsection{Comparison of Adversary Models}

\begin{figure}[h]
     \centering
     \includegraphics[width=\linewidth]{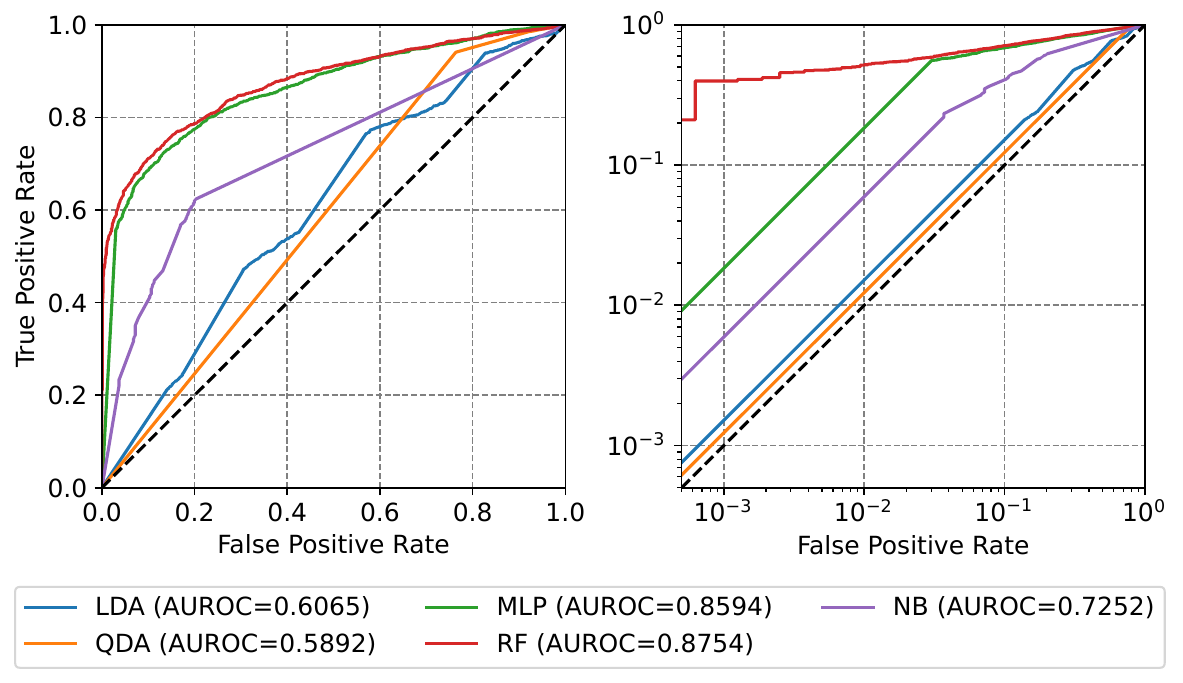}
     \vspace{-4mm}
    \caption{Comparison of AIA with different adversarial models and a batch size of $1$ on the Adult dataset.}
    \label{fig:adv_model_roc}
    \vspace{-3mm}
\end{figure}

We conduct a preliminary experiment of attribute inference attack using five types of adversarial models, including Gaussian Naive Bayes (NB), Linear Discriminant Analysis (LDA), Quadratic Discriminant Analysis (QDA), Multi-layer Perceptron (MLP) with one hidden layer of $100$ neurons, and Random Forest (RF) with $50$ estimators on the Adult dataset with $1{,}000$ shadow data samples and a batch size of $1$. Figure~\ref{fig:adv_model_roc} plots the ROC curves on both linear and log scales. We observe that the ensemble learning method (Random Forest) performs the best, especially in the low FPR region.

\subsection{Full Results From Main Paper}

\begin{figure*}[h]
     \centering
     \begin{subfigure}[b]{0.41\textwidth}
         \centering
         \includegraphics[width=\textwidth]{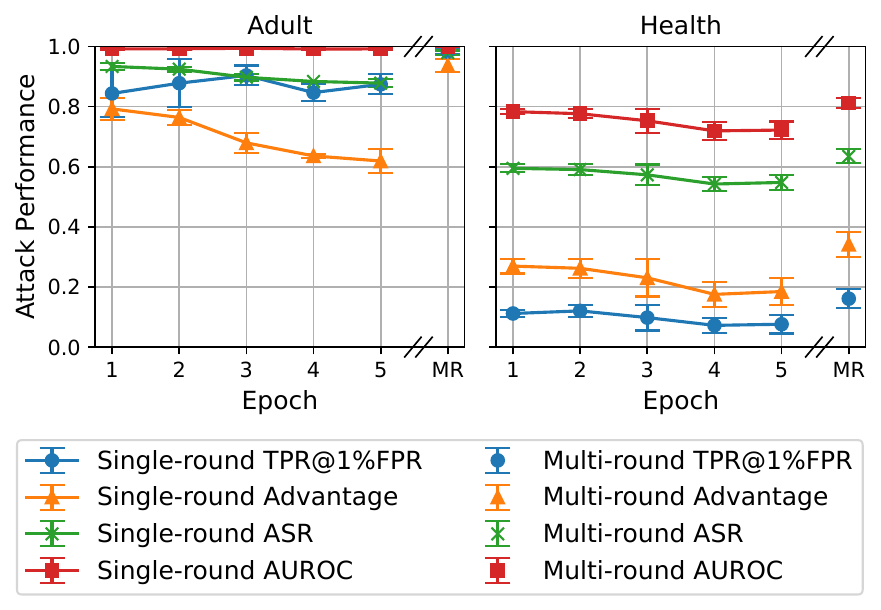}
         \caption{\textit{Attribute Inference Attack} with a batch size of 16.}
         \label{fig:sr_mr_aia}
     \end{subfigure}
     \hspace{14mm}
     \begin{subfigure}[b]{0.41\textwidth}
         \centering
         \includegraphics[width=\textwidth]{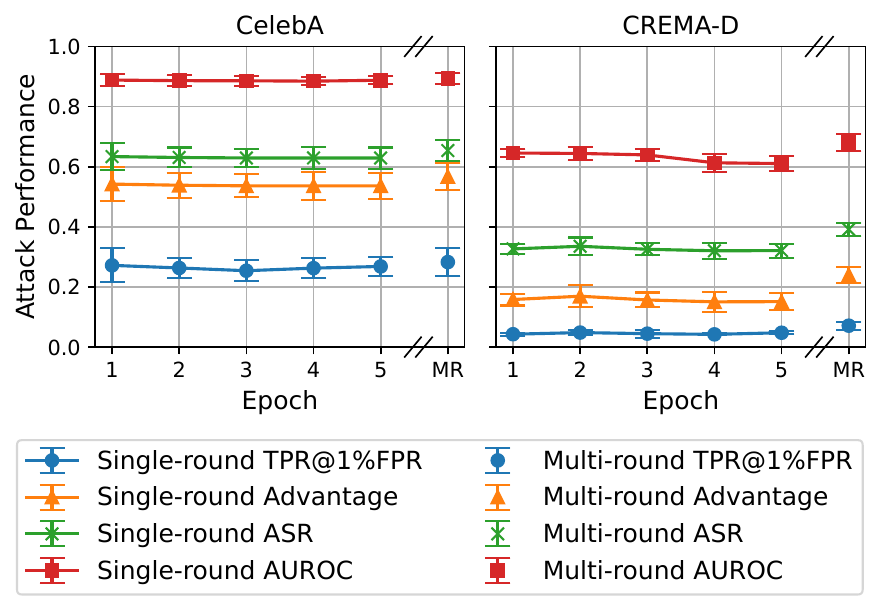}
         \caption{\textit{User Inference Attack} with a batch size of 8.}
         \label{fig:sr_mr_uia}
     \end{subfigure}

     \begin{subfigure}[b]{\textwidth}
         \centering
         \includegraphics[width=\textwidth]{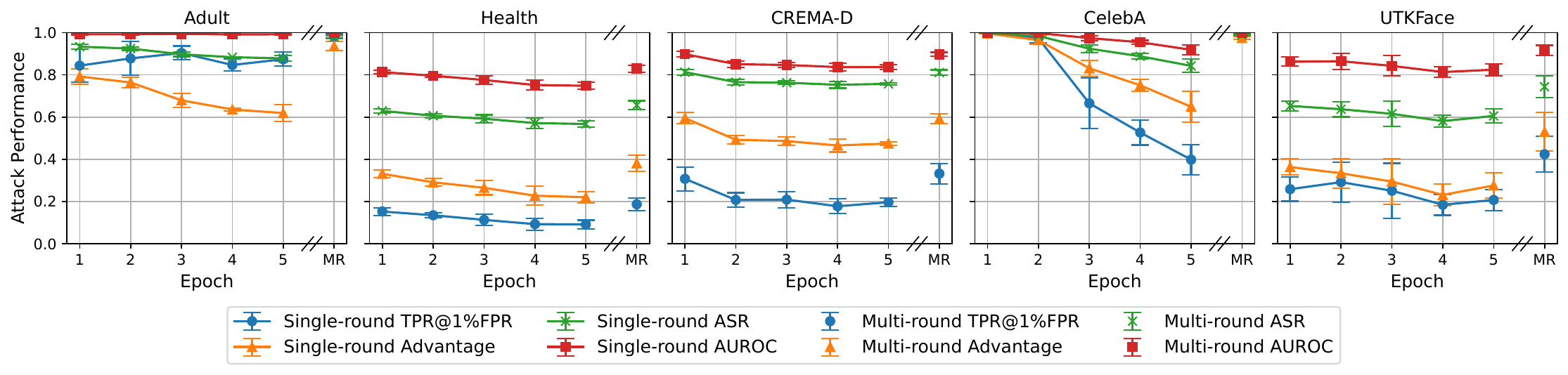}
         \caption{\textit{Property Inference Attack} with a batch size of 16.}
         \label{fig:sr_mr_pia}
     \end{subfigure}

     \begin{subfigure}[b]{\textwidth}
         \centering
         \includegraphics[width=\textwidth]{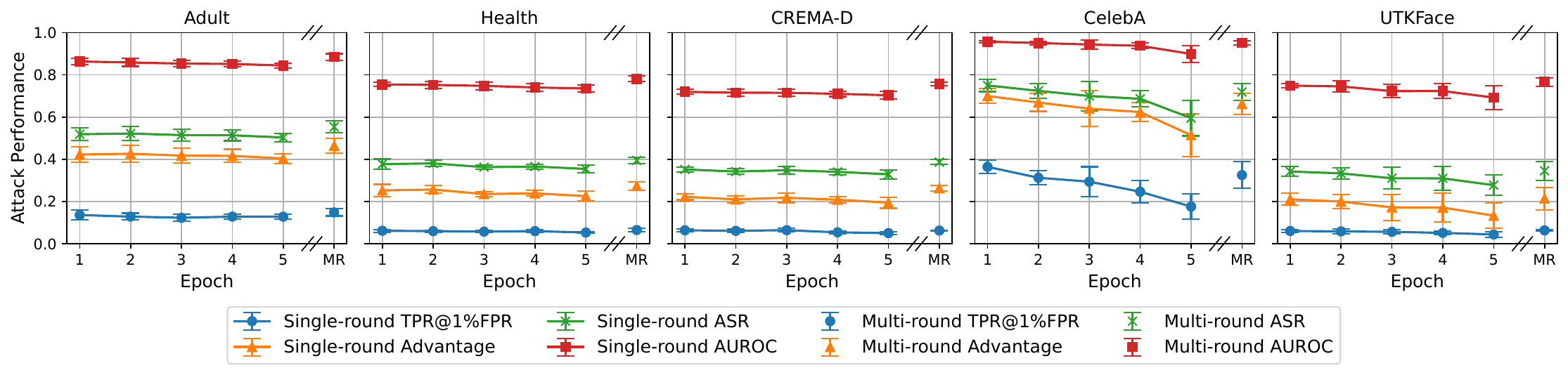}
         \caption{\textit{Distributional Inference Attack} with a batch size of 128.}
         \label{fig:sr_mr_dia}
     \end{subfigure}

    \caption{Comparison of single-round and multi-round inference attacks.}
    \label{fig:sr_mr}
\end{figure*}

\begin{figure*}[h]
     \centering
     \begin{subfigure}[b]{\textwidth}
         \centering
         \includegraphics[width=\textwidth]{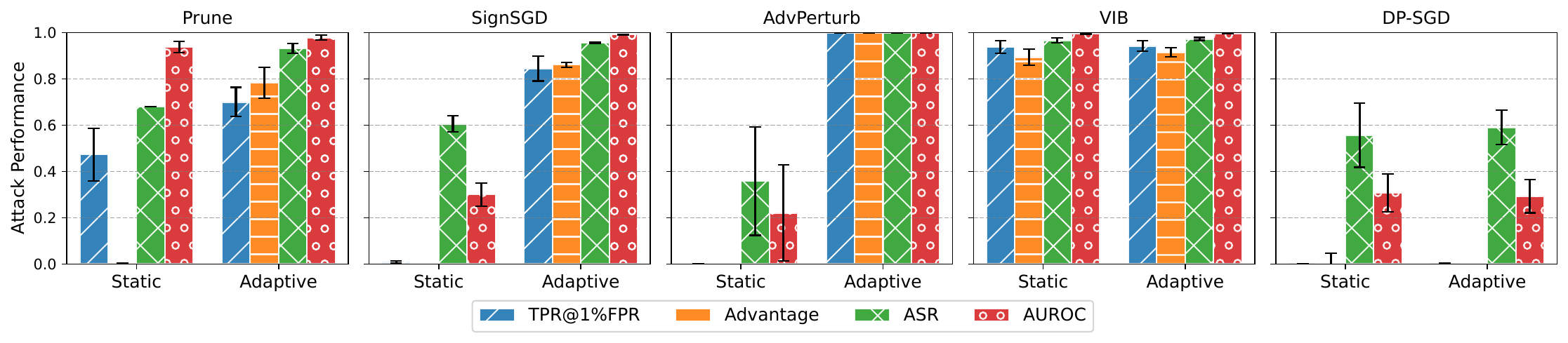}
         \caption{\textit{Attribute Inference Attack} on the Adult dataset with a batch size of 16.}
         \label{fig:aia_def_adult}
     \end{subfigure}

     \begin{subfigure}[b]{\textwidth}
         \centering
         \includegraphics[width=\textwidth]{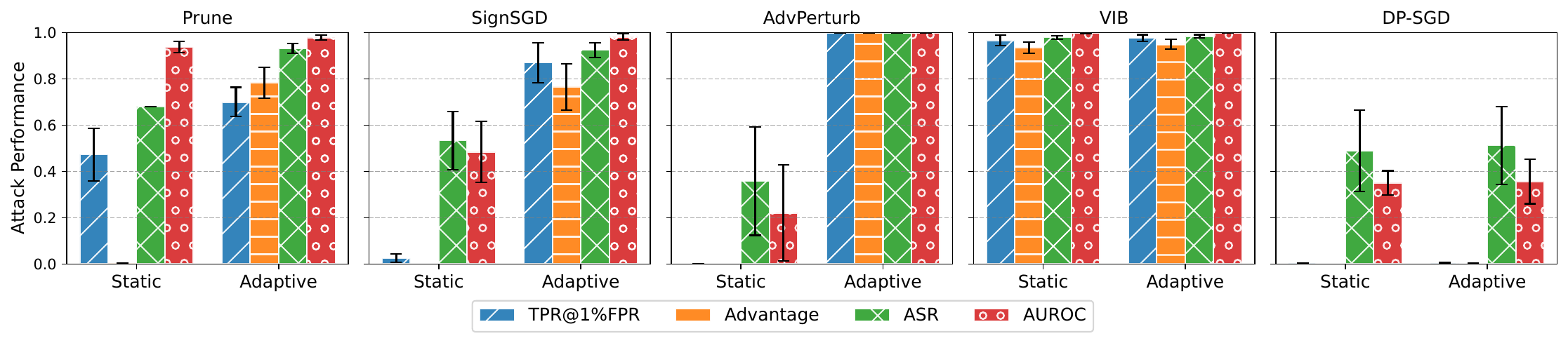}
         \caption{\textit{Property Inference Attack} on the Adult dataset with a batch size of 16.}
         \label{fig:pia_def_adult}
     \end{subfigure}

     \begin{subfigure}[b]{\textwidth}
         \centering
         \includegraphics[width=\textwidth]{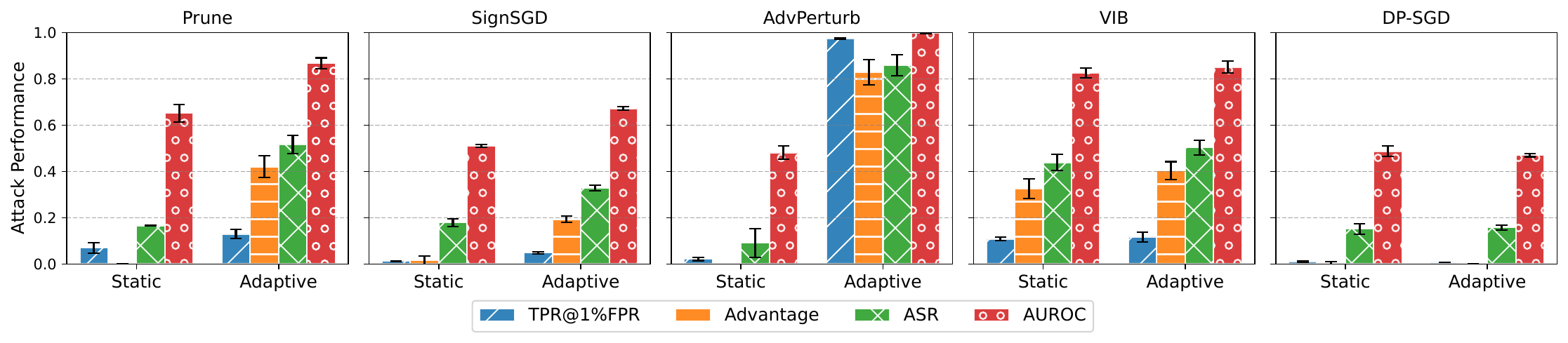}
         \caption{\textit{Distributional Inference Attack} on the Adult dataset with a batch size of 128.}
         \label{fig:dia_def_adult}
     \end{subfigure}

     \begin{subfigure}[b]{\textwidth}
         \centering
         \includegraphics[width=\textwidth]{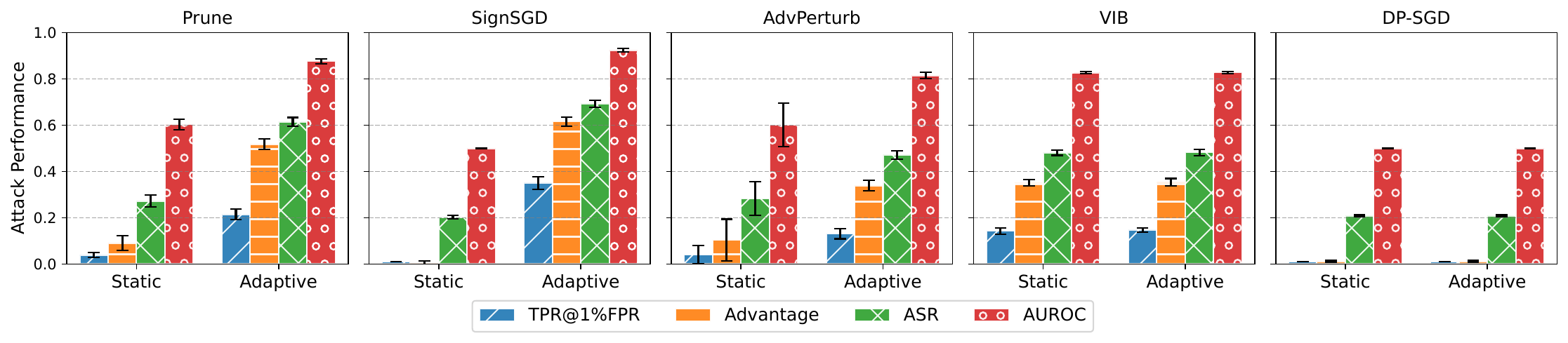}
        \caption{\textit{User Inference Attack} on the CelebA dataset with a batch size of 8.}
        \label{fig:uia_def_celeba}
     \end{subfigure}

    \caption{Comparison of various defenses against static and adaptive adversaries.}
    \label{fig:defenses_full}
\end{figure*}

Figure~\ref{fig:sr_mr} provides a complete comparison between single-round and multi-round inference attacks on all datasets. Figure~\ref{fig:defenses_full} evaluates defense against all four types of inference attacks under both the static and adaptive adversary settings.

\subsection{\revise{Impact of Model Size}}

\begin{figure*}[t]
    \centering
    \includegraphics[width=0.94\textwidth]{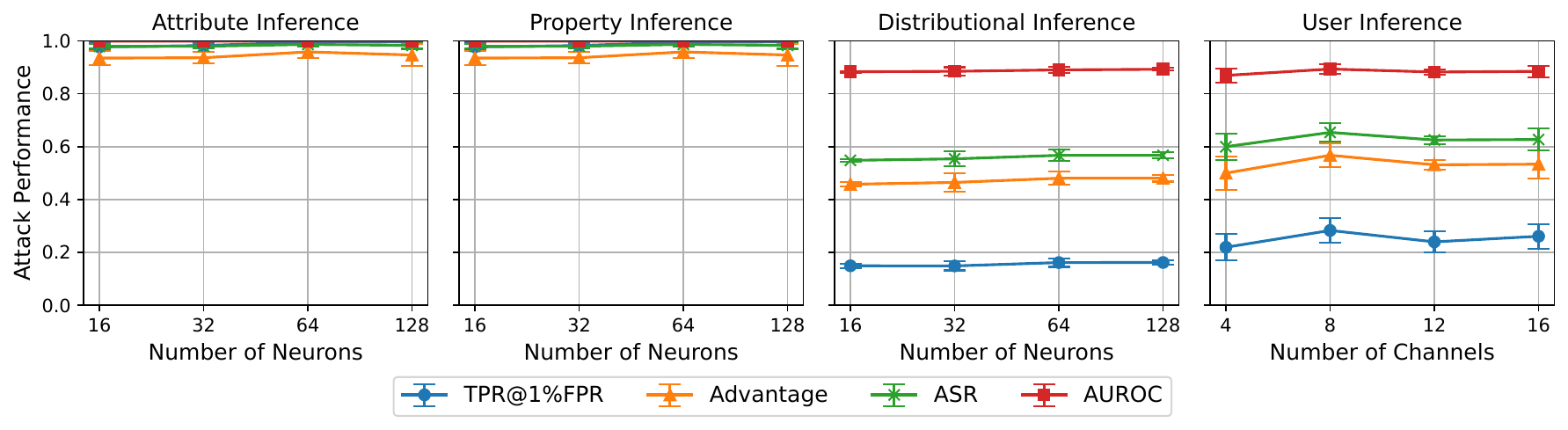}
    \caption{\revise{Sensitivity analysis of the impact of varying model sizes on the performance of inference attacks on the Adult (AIA, PIA, DIA) and CelebA (UIA) datasets.}}
    \label{fig:model_size}
\end{figure*}

\revise{We conduct additional experiments to study the impact of model size for all four types of inference attacks in Figure~\ref{fig:model_size}.}

\end{document}